\documentclass[onecolumn]{IEEEtran}
\pdfoutput=1

\usepackage{float}
\restylefloat{table}
\usepackage{pseudocode}
\usepackage{rotating}
\usepackage{algorithmicx}
\usepackage{epsfig}
\usepackage{caption}
\usepackage{pdfpages}
\usepackage{graphicx}
\usepackage{multirow}
\usepackage{amsmath,graphicx}
\usepackage{amssymb}
\usepackage{epstopdf}
\usepackage{subfigure}
\usepackage{color}
\usepackage{amsthm}

\theoremstyle{definition}

\theoremstyle{remark}

\usepackage{array}
\usepackage{booktabs}
\newcommand{\bkt}[1]{\left( {#1} \right)}
\newcommand{\brac}[1]{ \{ #1 \} }
\newcommand{\inner}[1]{\langle {#1} \rangle }
\newcommand{\abs}[1]{| {#1} | }
\newcommand{\mbf}[1]{\mathbf{{#1}}  }

\begin{document}
\title{Iterative non-local shrinkage algorithm for undersampled MR image reconstruction}
\author{Yasir Q. Mohsin, Gregory Ongie, and Mathews~Jacob,~\IEEEmembership{Senior Member,~IEEE}
\thanks{ Yasir Q. Mohsin and Mathews Jacob are with the Department of Electrical and Computer Engineering, Univ. Iowa, IA, USA. Gregory Ongie is with the Department of Mathematics, Univ. Iowa, IA, USA (emails: \{yasir-mohsin,gregory-ongie,mathews-jacob\}@uiowa.edu).
\newline This work is supported by grants  NSF CCF-0844812, NSF CCF-1116067, NIH 1R21HL109710-01A1, ACS RSG-11-267-01-CCE, and ONR grant N00014-13-1-0202.}}

\maketitle

\begin{abstract}
\emph{We introduce a fast iterative non-local shrinkage algorithm to recover MRI data from undersampled Fourier measurements. This approach is enabled by the reformulation of current non-local schemes as an alternating algorithm to minimize a global criterion. The proposed algorithm alternates between a non-local shrinkage step and a quadratic subproblem. We derive analytical shrinkage rules for several penalties that are relevant in non-local regularization. The redundancy in the searches used to evaluate the shrinkage steps are exploited using filtering operations. The resulting algorithm is observed to be considerably faster than current alternating nonlocal algorithms. The comparisons of the proposed scheme with state-of-the-art regularization schemes show a considerable reduction in alias artifacts and preservation of edges. }
\end{abstract}


\section{Introduction}
\label{sec:intro}

Non-local means (NLM) denoising algorithms were originally introduced to exploit the similarity between patches in an image to suppress noise \cite{buades2006denoising,uinta,buades2006review}. These methods recover each pixel in the denoised image as a weighted linear combination of all the pixels in the noisy image; the weights between any two pixels were estimated from the noisy image as the measure of similarity between their patch neighborhoods. This algorithm has been extended to deblurring problems by reformulating it as a regularized reconstruction scheme, where the regularization penalty is the weighted sum of square differences between all the pixel pairs in the image \cite{cohen2008non,gilboa2006nonlocal,lou2010image}. The weights are estimated from the noisy or blurred images itself, similar to classical NLM schemes. One of the difficulties in applying this scheme to challenging inverse problems (e.g. MRI recovery from under sampled data) is the dependence of the criterion on pre-specified weights; the use of the weights estimated from aliased images often preserve the alias patterns rather than suppressing them. Some authors have shown that iterating between the denoising and weight estimation step improves the quality of the images in deblurring applications \cite{Peyre}, but often had limited success in heavily undersampled Fourier inversion problems.

The alternating NLM scheme has been recently shown to be a majorize-minimize algorithm to solve for a penalized optimization problem; the penalty term is the sum of unweighted robust distances between image patches \cite{Wendy,wang2012penalized,yangisbi11}. The above reinterpretation was motivated by half quadratic
regularization methods used in the context of pixel-based smoothness regularization \cite{Geman:1995p9732,charbonnier1997deterministic,delaney1998globally,nikolova2001fast}.
The quality of the images recovered using the resulting NLM methods are heavily dependent on the specific distance metric used for inter-patch comparisons. While convex metrics such as $\ell_{1}$ distances may be used, nonconvex metrics that correspond to the classical NLM choices are seen to provide significantly improved results \cite{Wendy}. Since the direct alternation between weight estimation and optimization are not guaranteed to converge to the global minimum when nonconvex metrics are used, continuation strategies are utilized to minimize the convergence of the algorithm to local minima \cite{Wendy}. The main challenge associated with the implementation in \cite{Wendy} is the high computational complexity of the alternating minimization algorithm.

In this paper, we introduce a novel iterative algorithm to directly minimize the robust non-local criterion. This approach is based on a quadratic majorization of the patch based penalty term. Unlike the majorization used in our previous work, the weights of the quadratic terms are identical for all patch pairs, but now involves a new auxiliary variable.  Similar half-quadratic strategies are widely used in the context of sparse optimization \cite{Geman:1995p9732,charbonnier1997deterministic,delaney1998globally,nikolova2001fast}. The proposed algorithm alternates between two main steps (\textbf{a}) non-local shrinkage to determine the auxiliary variable, and (\textbf{b}) a quadratic optimization problem. We re-express the quadratic penalty involving the sum of patch differences as one involving sum of pixel differences, which enables us to solve for the quadratic sub-problem analytically. We derive analytical shrinkage expressions for a range of distance functions that are relevant for non-local regularization; this generalizes the shrinkage formulae derived by Chartand in the context of $\ell_{p}$ penalties \cite{chartrand2007exact}. Note that each step of the iterative shrinkage algorithm is fundamentally different from classical non-local schemes that solve an weighted quadratic optimization at each step \cite{buades2006denoising,uinta,buades2006review,cohen2008non,gilboa2006nonlocal,lou2010image}. The direct evaluation of the shrinkages of the patches is computationally expensive. We propose to exploit the redundancies in the shrinkages at adjacent pixels using separable filtering operations, thereby considerably reducing the computational complexity.

We compare the convergence of the proposed scheme with the iterative reweighted algorithm in our previous implementation \cite{Wendy}. We observe that the proposed scheme is approximately seven times faster than our previous iterative reweighted formulation. We also compared several distance functions in the context of iterative non-local shrinkage algorithm.  Our comparisons show that saturation of the distance function is key to good performance in non-local algorithms since each patch is compared with several other patches. The saturation is needed in non-local schemes unlike local TV methods, where a specified pixel is only compared with its neighbors. Our comparisons show that the truncated $\ell_{p}; p=0.5$ penalty provides the best results. We perform extensive comparisons of the scheme against local total variation (TV) regularization and a recent dictionary learning algorithm, which also exploits the similarity between image patches. The experiments demonstrates the considerable benefits in using non-local regularization. \label{sec:intro}

The rest of this paper is organized as fellows. We briefly describe the background in Section II. The proposed iterative non-local shrinkage algorithm is detailed in Section III, while the details of the implementation is outlined in Section IV. Section V demonstrates the performance of our method on numerous examples using CS and denoising techniques.
\section{Background}

\subsection{Unified Non-Local Formulation}
The iterative algorithm that alternates between classical non-local image recovery \cite{osher} and the re-estimation of weights was shown \cite{Wendy} to be a majorize-minimize (MM) algorithm to solve for
\begin{equation}
\label{cost}
\widehat{\mbf f} = \arg \min_{\mbf f} \underbrace{\|\mathbf A\mathbf f-\mathbf b\|^{2}+ \lambda \mathcal{G}(\mathbf f)}_{{\cal C}(\mathbf f)},
\end{equation}
where $\mbf f \in \mathbb C^{N}$ is a vector obtained by the concatenating the rows in a 2-D image $f(\mbf x), \mbf x \in \mathbb Z^{2}$; $\mathbf A \in \mathbb C^{M\times N}$ is a matrix that models the measurement process; and $\mbf b \in \mathbb C^{M}$ is the vector of measurements. While the first term in the cost function enforces data fidelity in k-space, the second term enforces sparsity. The regularization functional $\mathcal{G}(\mbf f)$ is specified by:
\begin{equation}
\label{Reg}
\mathcal{G}(\mbf f)=\sum_{\mathbf x}\sum_{\mathbf y\in \mbf x + {\cal N}}\varphi\left(P_{\mathbf{x}}(\mbf f)-P_{\mathbf{y}}(\mbf f)\right).
\end{equation}
Here, $\varphi$ is an appropriately chosen potential function and $P_{\mbf x}$ is a patch extraction operator which extracts an image patch centered at the spatial location $\mbf x$ from the image $\mbf f$:
\begin{equation}
P_{\mbf x}(\mbf f) = f(\mbf x + \mbf p),\quad \mbf p \in \cal B,
\end{equation}
where $\cal B$ denotes the indices in the patch. For example, if we choose a square patch of size $(2N +1)$, the set ${\cal B} = [-N,...,N]\times [-N,...,N]$. Similarly, $\cal N$ are the indices of the search neighborhood; the patch $P_{\mbf x}(\mbf f)$ is compared to all the patches whose centers are specified by $\mbf x+\cal N$. For example, if we choose a square shaped neighborhood of size $2M+1$, the set ${\cal N} = [-M,...,M]\times [-M,...,M]$. The shape of the patches and the search neighborhood may be easily changed by re-defining the sets $\cal N$ and $\cal B$.

In this paper, we focus on potential functions of the form
\begin{equation}
\label{specialcase}
\varphi(\mbf g) = \phi\bkt{\|\mbf g\|},
\end{equation}
where $\phi: \mathbb R^{+}\rightarrow \mathbb R^{+}$ is an appropriately chosen distance metric and $\|\mathbf g\|^{2}=\sum_{\mathbf p \in \cal B} |g(\mbf p)|^{2}$. However, the theory presented in this paper is general enough to work for other potential functions.

\subsection{Solution Using Iterative Reweighted Algorithm}
We showed in \cite{Wendy} that (\ref{cost}) can be solved using MM scheme, where the regularization term is majorized as the weighted sum of patch differences:
\begin{equation}
\label{major}
\mathcal{G}(\mbf f)\leq\underbrace{\sum_{\mathbf x}\sum_{\mathbf p\in {\cal N}}w_{n}(\mathbf{x},\mbf x + \mbf p)
\|P_{\mathbf{x}}(\mbf f)-P_{\mathbf{x+p}}(\mbf f)\|^2}_{\mathcal{G}_{n}(\mbf f)}.
\end{equation}
The weights are specified by:
\begin{equation}
\label{weights}
w_{n}(\mathbf{x},\mathbf{y})=\frac{\phi'
(\|P_{\mathbf{x}}(\mbf f_{n})-P_{\mathbf{y}}(\mbf f_{n})\|)}
{2\|P_{\mathbf{x}}(\mbf f_{n})-P_{\mathbf{y}}(\mbf f_{n})\|}.
\end{equation}
Here, $\mbf f_{n}$ is the function at the $n^{\rm th}$ iteration. Each iteration of the MM algorithm involves the minimization of the criterion
\begin{equation}
\label{newcost}
\mbf f_{n+1} = \arg \min_{\mbf f} \|\mbf A\mbf f-\mbf b\|^{2}+ \lambda\, \mathcal{G}_{n}(\mbf f).
\end{equation}
Note that this optimization problem is essentially the classical non-local $H_{1}$ regularization scheme \cite{osher}. The alternation between (\ref{major}) and the re-computation of the weights (\ref{weights}) will converge to the local minimum of the criterion (\ref{cost}). Continuation strategies were used in \cite{Wendy} to improve the convergence of the algorithm to the global minimum of (\ref{cost}). As discussed above, one of the main challenges of the algorithm is its high computational complexity. Specifically, the conjugate gradients algorithm to solve the quadratic sub-problem converges slowly as the value of the weights increase.
\section{Proposed Algorithm}
\subsection{Majorization of the Penalty Term}
In this work, we will consider an alternate majorization of the potential function $\varphi$ specified by
\begin{equation}
\label{new}
\varphi(\mbf t) = \min_{\mbf s} \left\{\psi(\mbf s) + \frac{\beta}{2}\|\mbf s-\mbf t\|^{2} \right\}.
\end{equation}
Here, $\mbf s$ is an auxiliary variable, $\beta > 0$ is an arbitrarily chosen scalar parameter, and $\psi$ is a function that is dependent on $\varphi$ and $\beta$. Using (\ref{new}), we majorize the cost function $\cal C$ in (\ref{cost}):
\begin{equation}
\label{newcost_hq}
{\cal C}(\mbf f) = \min_{\brac{\mbf s_{\mbf x,\mbf q}}} \|\mbf A\mbf f-\mbf b\|^{2} + \lambda \frac{\beta}{2}\sum_{\mbf x,\mbf q}\|P_{\mathbf{\mbf x}}(\mbf f)-
P_{\mathbf{x+p}}(\mbf f)-\mbf s_{\mbf x,\mbf q}\|^{2} +\lambda\sum_{\mbf x}\sum_{\mbf q\in \cal N}\psi(\mbf s_{\mbf x,\mbf q})
\end{equation}
We use an alternating minimization algorithm to optimize \eqref{newcost_hq}. Specifically, we alternate between the determination of the optimal variables $\brac{\mbf s_{\mbf x,\mbf q}}$, assuming $\mbf f$ to be fixed and the determination of the optimal $\mbf f$, assuming $\brac{\mbf s_{\mbf x,\mbf q}}$ to be fixed. The convergence of the above scheme could potentially be improved by including augmented Lagrangian terms \cite{wu2010augmented} or Bregman iterations \cite{goldstein2009split}. However, the use of these methods often interfered with the continuation strategies used to improve the convergence of the algorithm to the global minima; these schemes were designed for convex cost functions, where local minima issues do not exist.

\subsection{The $\mbf s$ Sub-Problem: solve for $\mbf s_{\mbf x,\mbf q}$, assuming $\mbf f$ fixed}
If the variable $\mbf f$ is assumed to be a constant, the determination of each of the auxiliary variables $\mbf s_{\mbf x,\mbf q}$ corresponding to different values of $\mbf x$ and $\mbf y$ can be treated independently:
\begin{equation}
\label{newcost}
\bar{\mbf s}_{\mathbf x,\mbf q} = \arg \min_{\mbf s_{\mbf x,\mbf q}}\frac{\beta}{2}\|P_{\mathbf{\mbf x}}(\mbf f)-
P_{\mathbf{x+q}}(\mbf f)-\mbf s_{\mbf x,\mbf q}\|^{2}\\ + \psi(\mbf s_{\mbf x,\mbf q}).
\end{equation}
We will show in the subsection \ref{majsec} (see (\ref{combined})) that $\mbf s_{\mbf x,\mbf q}$ can be determined analytically as a shrinkage for all penalties of interest
\begin{equation}
\label{pshrinkage}
\bar{\mbf s}_{\mbf x,\mbf q} =\left[P_{\mbf x}(\mbf f)-P_{\mbf x + \mbf q}(\mbf f)\right]~\nu\bkt{\|P_{\mbf x}(\mbf f)-P_{\mbf x + \mbf q}(\mbf f)\|},
\end{equation}
where $\nu: \mathbb R^{+} \rightarrow \mathbb R^{+}$ is a function that is dependent on the distance metric $\phi$.

Note that the structure of the algorithm is exactly the same for different choices of distance function;  the analytical expressions for the shrinkage steps will change depending on the specific choice. We will determine the shrinkage rules corresponding to the useful penalties in the next section.
\begin{table*}[!t]
\vspace{-7mm}
\caption{Distances functions $\phi(t)$ that are relevant in non-local regularization (first row) and the associated shrinkage rules $t\cdot\nu(|t|)$ (second row); see Appendix B for the corresponding formulas. Here we illustrate the shrinkage rules in 1-D for the parameter choices $\beta = 2$, $p=0.5$, $T=1$, and $\sigma = 0.5$.
The approach introduced in the paper enables the evaluation of shrinkage rules for a much larger class of penalties, generalizing the results in \cite{chartrand2007exact} for $\ell_{p}$ penalties shown in the first column. }
\label{table:shrinkage}
\centering
\def \smallfigwidth{0.18\textwidth}
\begin{tabular}{ c c c c c }
\small $\ell_{p}$ & \small $\ell_{p}$-$T$ & \small $H_1$ & \small Peyre & \small NLTV\\
\includegraphics[height=\smallfigwidth]{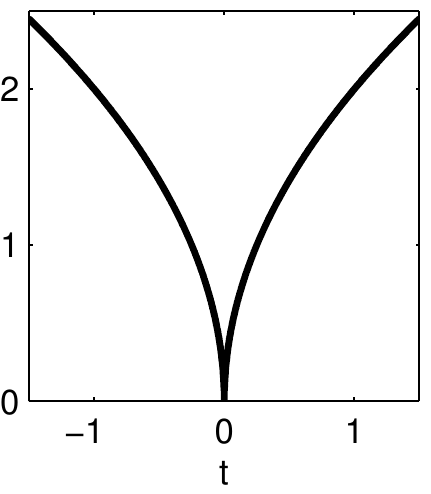}\vspace{0.05cm} &
\includegraphics[height=\smallfigwidth]{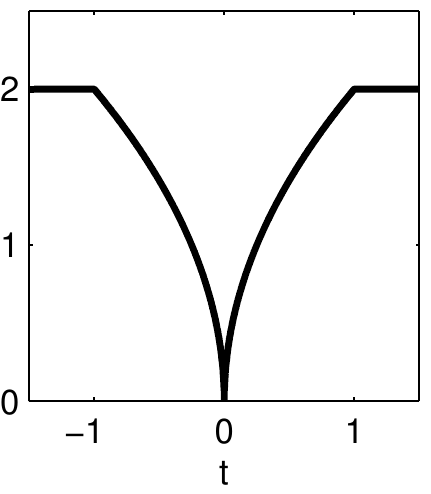}\vspace{0.05cm} &
\includegraphics[height=\smallfigwidth]{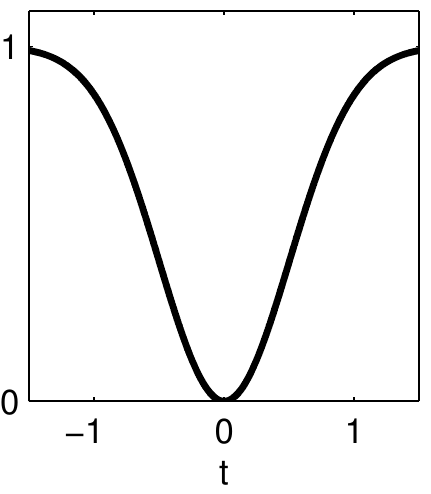}\vspace{0.05cm} &
\includegraphics[height=\smallfigwidth]{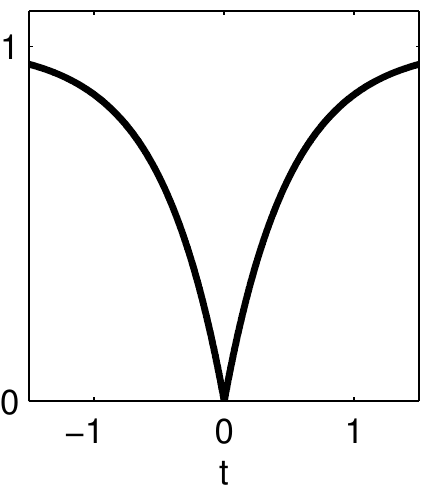}\vspace{0.05cm} &
\includegraphics[height=\smallfigwidth]{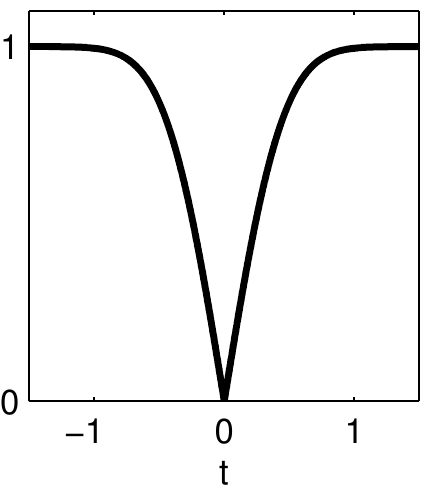}\vspace{0.05cm} \\
\includegraphics[height=\smallfigwidth]{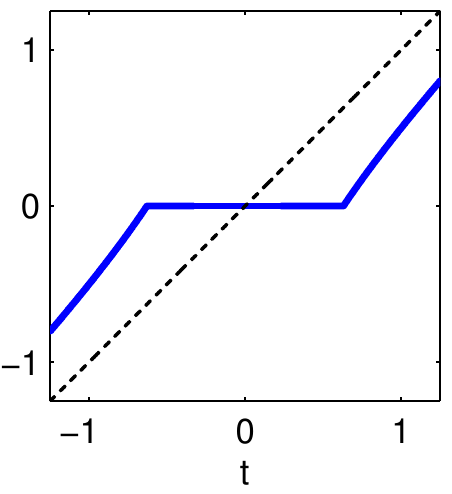} &
\includegraphics[height=\smallfigwidth]{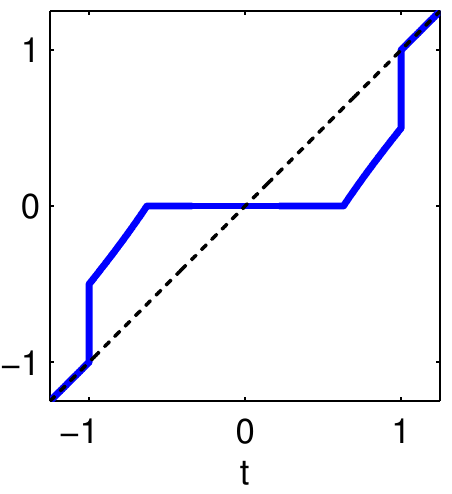} &
\includegraphics[height=\smallfigwidth]{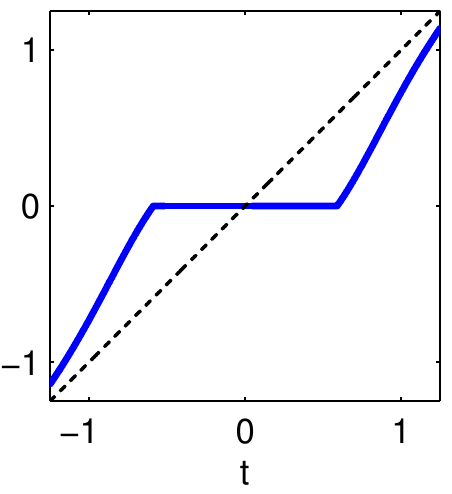} &
\includegraphics[height=\smallfigwidth]{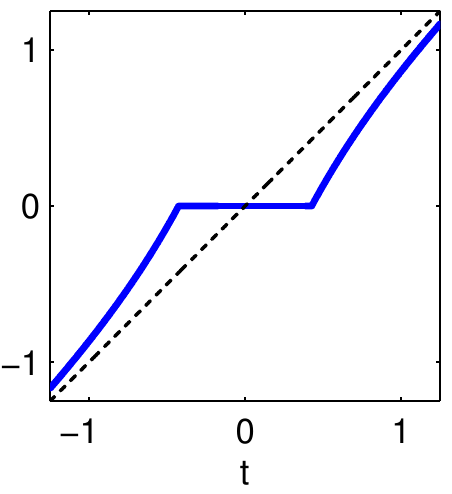} &
\includegraphics[height=\smallfigwidth]{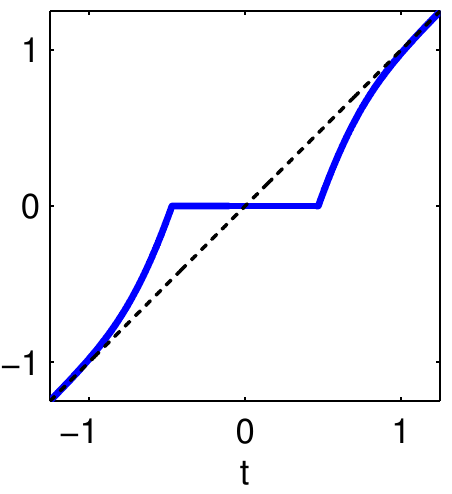} \\
 \end{tabular}
\end{table*}

\subsection{Determination of Shrinkage Rules}
\label{majsec}
Several potential functions are currently available \cite{Wendy}, resulting in different flavors of robust non-local regularization. The shrinkage rules for the cases for $\ell_{p}$ penalties with $p\leq 1$ are available in \cite{chartrand2007exact}. We now determine the corresponding shrinkage rules for a larger class of non-local penalties.

\begin{figure}[ht!]
 \centering
\subfigure[$\varphi(t)$]{\hspace{-0.5em}\includegraphics[width=0.15\textwidth, height=0.115\textwidth]{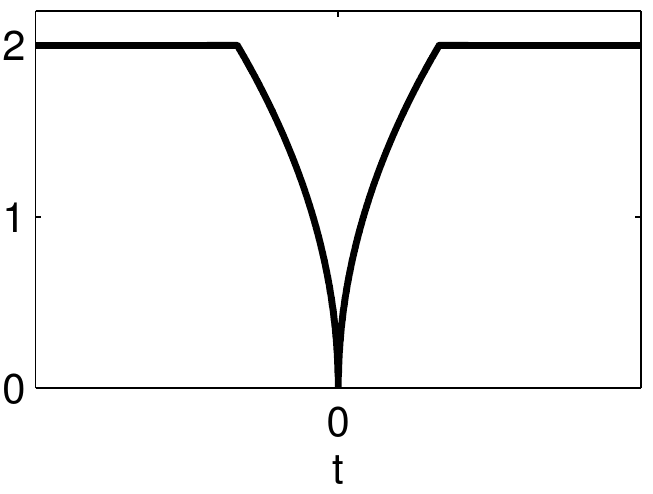}}\hspace{0.5em}
\subfigure[$r(t)$ and $\hat{r}(t)$]{\hspace{-0.5em}\includegraphics[width= 0.15\textwidth,height=0.115\textwidth]{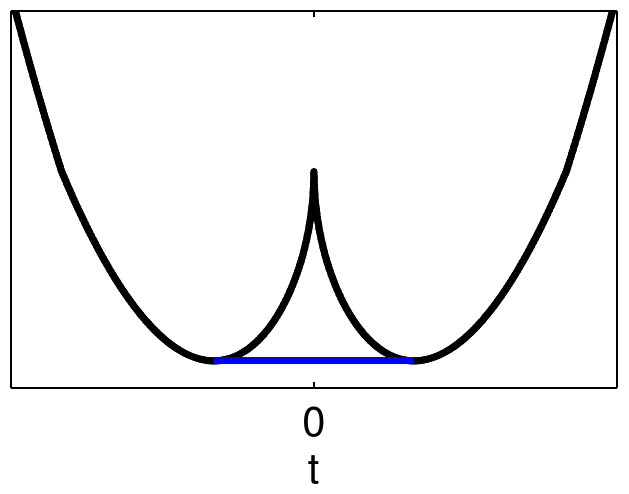}}\hspace{0.5em}
\subfigure[$\hat{\varphi}(t)$]{\hspace{-0.5em}\includegraphics[width= 0.15\textwidth,height=0.115\textwidth]{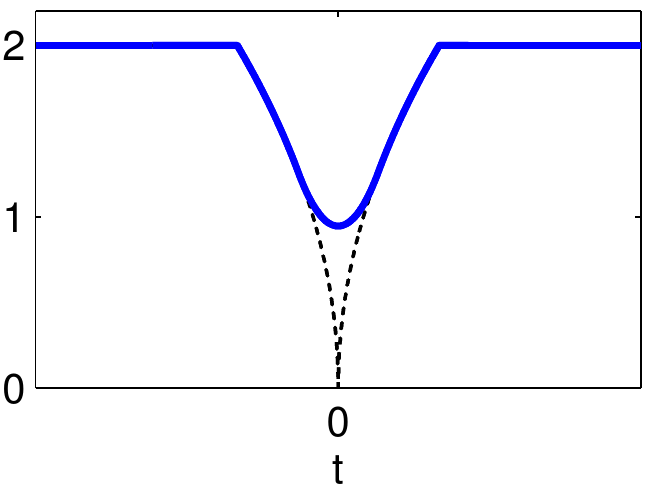}}\hspace{0.5em}\\
\subfigure[$\hat{\varphi}(t)$ for various $\beta$]{\includegraphics[height= 0.2\textwidth]{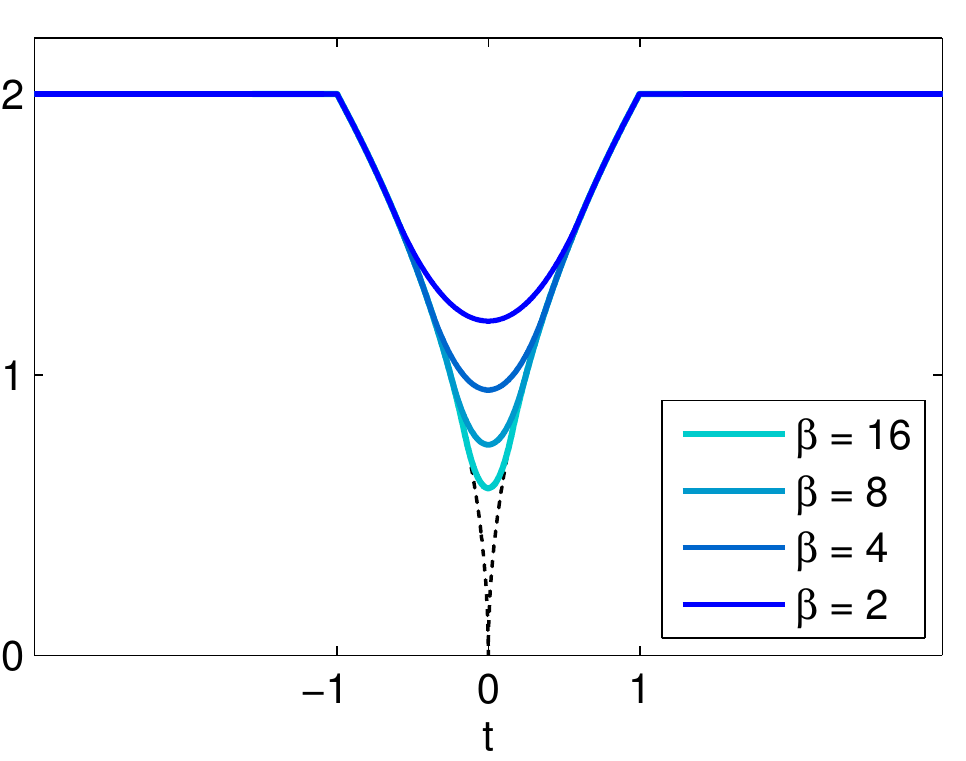}}\hspace{0.5em}
\subfigure[$t\cdot\nu(|t|)$ for various $\beta$]{\hspace{-0.5em}\includegraphics[height=0.20\textwidth]{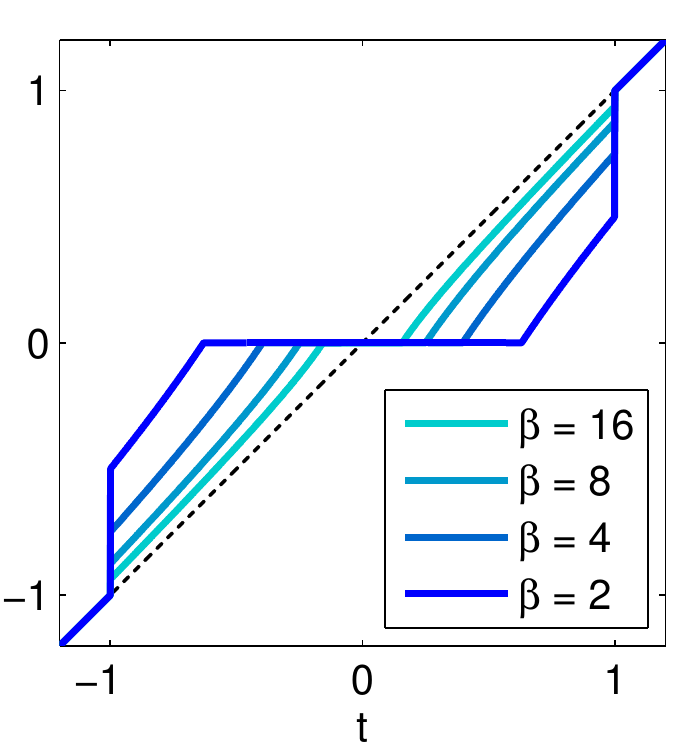}}\hspace{0.5em}
\caption{Approximation of the potential function: (a) shows the original potential function $\varphi(t)$ in 1-D, which is the truncated $\ell_{p}$; $p=0.5$ penalty, $T=1$. (b) indicates the corresponding ${r(t)= t^{2}/2-\frac{1}{\beta}\varphi(t)}$ function with $\beta =2$, shown in black. Note that this function is non-convex. Hence, we approximate this function by $\hat r(t)$ shown in blue, which is the best convex approximation of $r(t)$. The corresponding modified potential function is shown in blue in (c). (d) indicates the approximations for different values of $\beta$. Note that the approximations converge uniformly to $\varphi$. (e) shows the corresponding shrinkage rules. The potental functions and shrinkage rules for different penalties are shown in Table \ref{table:shrinkage}.}
\label{approximation}
\end{figure}

The majorization rule in (\ref{new}) can be rewritten as:
\begin{equation}
\label{simplification}
\underbrace{\frac{\|\mbf t\|^{2}}{2} - \frac{1}{\beta}~\varphi(\mbf t)}_{r(\mbf t)} = \max_{\mbf s}\left\{\inner{\mbf s, \mbf t} - \underbrace{\bkt{\frac{1}{\beta}~\psi(\mbf s) + \frac{\|\mbf s\|^{2}}{2}}}_{g(\mbf s)}  \right\}
\end{equation}
From the theory in \cite{convexdual}, the above majorization relation is satisfied when $r$ is a convex function, in which case $g = r^{*}$, the Legendre-Fenchel dual (or convex conjugate) of $r$:
\begin{equation}
{r^{*}(\mbf s)} = \max_{\mbf t}\left\{\inner{\mbf s, \mbf t} - r(\mbf t)  \right\}.
\end{equation}
However, the function $r$ is not convex for most penalties $\varphi$ that we are interested in, especially for small values of $\beta$. When $r$ is not convex, we propose to approximate $r$ by a convex function $\hat r$ so that relation \eqref{simplification} is satisfied.
We choose $\hat r$ such that the epigraph of $\hat r$ is the convex hull of the epigraph of $r$; $\hat r$ is thus the closest convex function to $r$; see Fig (\ref{approximation} b). For $\varphi$ functions of the form (\ref{specialcase}), we have $r(\mbf t) = q(\|\mbf t\|)$, where the function $q: \mathbb R^{+}\rightarrow \mathbb R^{+}$ is specified by $q(t) = t^{2}/2 - \phi(t)/\beta$.
In all the cases we consider in this paper (see Appendix B), we can obtain the convex hull approximation of $r$ as
\begin{equation}
\label{approximate}
\hat r(\mbf t) = \left\{ \begin{array}{ccc}
q(\|\mbf t\|) & \mbox{  if  } & q'(\|\mbf t\|)>0\\
c & \mbox{ else}
\end{array}\right.,
\end{equation}
where $c$ is an appropriately chosen constant to ensure continuity of $\hat r$; check Fig.\ (\ref{approximation}\! b).

The above convex hull approximation $\hat r$ of $r$ yields a cooresponding approximation $\hat \varphi$ of the potential function $\varphi$ given as
\[
	\hat \varphi(\mbf t) := \beta\left(\frac{\|\mbf t\|^{2}}{2} - \hat r(\mbf t)\right);
\]
see Fig.\ (\ref{approximation}\! c). For the potentials considered in this paper this results in the ``Huber-like'' approximation:
\[
	\hat \varphi(\mbf t) = \begin{cases}\frac{\beta\|\mbf t\|^2}{2}-c& \text{if}\quad \|\mbf t\| < L \\  \phi(\|\mbf t\|) & \text{else}\end{cases},
\]
 where $L = L(\beta) \rightarrow 0$ as $\beta\rightarrow\infty$. In particular we have $\hat \varphi \rightarrow \varphi$ uniformly as $\beta \rightarrow \infty$; see Fig.\ (\ref{approximation}\! d). Therefore, the following shrinkage rules can be interpreted as those corresponding to a Huber approximation of the potential $\varphi$, where the approximation improves with increasing $\beta$.

The shrinkage rule in (\ref{newcost}) involves the computation of $\bar{\mbf s}$ specified by:
\begin{align}\nonumber
\bar{\mbf s}(\mbf t) & = \arg \min_{\mbf s}\left\{\psi(\mbf s) + \frac{\beta}{2}\|\mbf s- \mbf t\|^{2} \right\}\\
			   & = \arg \max_{\mbf s}\{\,\inner{\mbf s,\mbf t} - \hat r^*(\mbf s)\,\},
\label{proxmap}
\end{align}
which is often called the proximal mapping of $\psi$. However, by exploiting duality we do not explicitly access to $\psi$ or $\hat r^*$ to compute $\bar{\mbf s}$. Differentiating the right hand side of \eqref{proxmap} with respect to $\mbf s$ and setting it to zero, we obtain $\mbf t - \partial \hat r^*(\bar{\mbf s}) \ni 0$, or equivalently, $\bar{\mbf s}(\mbf t) \in ({\partial \hat r^*})^{-1}( \mbf t )$.
Since the subgradients of Legendre-Fenchel duals satisfy $({\partial \hat r^*})^{-1}( \mbf t ) = ({\partial \hat r})( \mbf t )$, we have
\begin{equation}
\bar{\mbf s}(\mbf t) \in \partial \hat{r}(\mbf t).
\end{equation}
Considering the expression for the convex hull approximation of $\hat r$ in \eqref{approximate}, we have:
\begin{equation}\nonumber
\bar{\mbf s}(\mbf t)  =  \left\{\begin{array}{ccc}
q'(\|\mbf t\|)~\frac{\mbf t}{\|\mbf t\|}&\mbox{  if  }& q'(\|\mbf t\|)>0\\
0 & \mbox{else}.
\end{array} \right.
\end{equation}
Setting $q(t) = t^{2}/2 - \phi(t)/\beta$ in the above equation, we obtain the shrinkage rule as:
\begin{align}\nonumber
\label{combined}
\bar{\mbf s}(\mathbf t)  &=  \frac{\mathbf t}{\|\mathbf t\|} \bkt{\|\mbf t\| -  \frac{1}{\beta}~\phi'(\|\mbf t\|)}_{+}\\
&=  \mathbf t \underbrace{\bkt{1 - \frac{{\phi}'(\|\mbf t\|)}{\beta~\|\mbf t\|} }_{+}}_{\nu\bkt{\|\mbf t\|}},
\end{align}
where $(\cdot)_+ := \max\{\cdot, 0\}$. Here, $\nu(\|\mbf t\|)$ is a scalar between $0$ and $1$, which when multiplied by $\mbf t$ will yield the shrinkage of $\mbf t$. Setting $\mbf t = P_{\mbf x}(\mbf f) - P_{\mbf x+\mbf q}(\mbf f)$ in the above equation, we obtain the shrinkage rules to be used in (\ref{pshrinkage}). Note that the above approach can be adapted to most penalties.
We determine the shrinkage rules and the associated $\nu$ functions for common penalty functions $\varphi$ in non-local regularization in Appendix B. Table I shows the penalty functions for different metrics and the corresponding shrinkage weights.



\subsection{The $f$ Sub-Problem: solve for $f$, assuming $\mbf s_{\mbf x,\mbf q}$ fixed}
In this step, we assume the auxiliary variables $\mbf s_{\mbf x,\mbf q}$ to be fixed. Hence, the minimization of (\ref{cost}) simplifies to:
\begin{equation}
\label{newf}
\hat{\mbf f} = \arg \min_{\mbf f} \|\mbf A\mbf f-\mbf b\|^{2}+ \frac{\lambda\beta}{2}\underbrace{\sum_{\mathbf{x},\mathbf q\in {\cal N}}\|P_{\mathbf{x}}(\mbf f)-P_{\mathbf{x+q}}(\mbf f)-\mbf s_{\mbf x,\mbf q}\|^2}_{{\cal R}(\mbf f)}
\end{equation}
The above quadratic penalty may be solved using the conjugate gradients algorithm. However, we will now simplify it to an expression that can be solved analytically, which is considerably more efficient.

The quadratic penalty term involves differences between multiple patches in the image, each of which is a linear combination of quadratic differences between image pixels. The differences between two specific pixels are thus involved in different patch differences. We show in Appendix A that the pixel differences from several patches can be combined to obtain the following pixel-based penalty:
\begin{eqnarray}
\label{lastf}
\hat{\mbf f}=\arg \min_{\mbf f} \|\mbf A\mbf f-\mbf b\|^{2}+ \lambda\frac{\beta}{2}\sum_{\mathbf{q}\in{\cal N}} \|{\mbf D}_{\mathbf{q}}\mbf f-\mbf h_{\mathbf{q}}\|^{2}.
\end{eqnarray}
Here, ${\mbf D}_{\mathbf{q}}$ is the finite difference operator
\begin{equation}
({\mbf D}_{\mathbf{q}} \mbf f)(\mbf x) = \mbf f(\mathbf{x})-\mbf f(\mathbf{x}+\mathbf{q}).
\end{equation}
The images $\mbf h_{\mbf q}(\mbf x), \mbf q \in \mathcal N$, are obtained by shrinking the finite difference terms ${\mbf D}_{\mathbf{q}} \mbf f$:
\begin{equation}
\label{hq}
\mbf h_{\mbf q} =\bkt{{\mbf D}_{\mathbf{q}} \mbf f}\bullet \mbf v_{\mbf q},
\end{equation}
where $\bullet$ denotes the entrywise multiplication of the vectors, and the pixel shrinkage weights $\mbf v_{\mbf q}$ for a specified spatial location $\mbf x$ are obtained by the sum of the shrinkage weights for the nearby patch pairs
\begin{equation}
\label{wt}
\mbf v_{\mbf q}(\mbf x) = \sum_{\mathbf{p}\in{\cal B}}~\underbrace{\nu\bkt{\|P_{\mbf x+\mbf p}(\mbf f) - P_{\mbf x+\mbf p +\mbf q}(\mbf f)\|}}_{\mbf u_{\mbf q}(\mbf x)}.
\end{equation}
Here, $\nu$ is specified by (\ref{combined}). We solve (\ref{lastf}) in the Fourier domain for measurement operators $\mbf A$ that are diagonalizable in the Fourier domain, as shown in the next section.

\section{Implementation}
We now focus on the implementation of the sub-problems. Specifically, we show that all of the above steps can be solved analytically for most penalties and measurement operators of practical interest. This enables us to realize a computationally efficient algorithm. We also introduce a continuation scheme to improve the convergence of the algorithm.

\subsection{Analytical Solution of (\ref{newf}) in the Fourier Domain}
 The Euler-Lagrange equation for (\ref{lastf}) is given by:
\begin{equation}
\label{el}
 \left(2 \mbf A^{H}\mbf A  + \lambda \beta \sum_{\mbf q\in {\cal N}} {\mbf D}_{\mathbf{q}}^{H}{\mbf D}_{\mathbf{q}}\right) \mbf f = 2 \mbf A^{H}\mbf b  + \lambda \beta \sum_{\mbf q\in {\cal N}} {\mbf D}_{\mathbf{q}}^{H}\mbf h_{\mathbf{q}}
\end{equation}
Here $\mbf B^H$ denotes the Hermitian transpose of matrix $\mbf B$. Note that the variables in the left hand side of (\ref{el}) are fixed. Thus, this step involves the solution to a linear system of equations.
In many inverse problems of interest (e.g. Fourier sampling, deblurring), the measurement operator $\mbf A$ is diagonalizable in the Fourier domain, in which case we may write $\mbf A^H \mbf A$ as a pointwise multiplicaiton in the Fourier domain. For instance, in the particular case when $\mbf A$ is a Cartesian Fourier undersampling operator, we may write
\begin{equation}
	\mbf A^H\mbf A \mbf f= \mathcal{F}^{-1}(\mbf a \bullet \mathcal F(\mbf f))
\end{equation}
where $\mathcal{F}$ discrete Fourier transform and $\mbf a$ is a vector of ones and zeros corresponding to the Fourier sample locations. Likewise,
assuming circular boundary conditions for the finite difference operator $\mbf D_{\mbf q}$, the operator $\mbf D_{\mbf q}^H \mbf D_{\mbf q}$ is  diagonalizable in the Fourier domain as
\begin{equation}
	\mbf D_{\mbf q}^H\mbf D_{\mbf q} \mbf f = \mathcal{F}^{-1}(\mbf |\mbf d_{\mbf q}|^2 \bullet \mathcal F(\mbf f)),
\end{equation}
where $|\mbf d_{\mbf q}|^2$ is the pointwise modulus squared of the Fourier multiplier $\mbf d_{\mbf q}$ corresponding to $\mbf D_{\mbf q}$.
Hence, taking the DFT of both sides of (\ref{el}) we have
\begin{equation*}
 \left(2 \mbf a  + \lambda \beta \sum_{\mbf q\in {\cal N}} |\mbf d_{\mathbf{q}}|^2\right) \bullet \mathcal F(\mbf f) = 2 \mbf b_0  + \lambda \beta \mathcal{F}\left(\sum_{\mbf q\in {\cal N}} \mbf D_{\mathbf{q}}^H\mbf h_{\mathbf{q}}\right),
\end{equation*}
where $\mbf b_0 = \mathcal{F}(\mbf A^H \mbf b) \in \mathbb C^M$ is a zero-padded version of the Fourier samples $\mbf b \in \mathbb C^N$. Solving for $\mbf f$ gives
\begin{equation}
\label{f}
\mbf f = \mathcal F^{-1}\left[\frac{2 \mbf b_0  + \lambda \beta \mathcal{F}\left(\sum_{\mbf q\in {\cal N}} \mbf D_{\mathbf{q}}^H\mbf h_{\mathbf{q}}\right)}{2 \mbf a  + \lambda \beta \sum_{\mbf q\in {\cal N}} |{\mbf d}_{\mathbf{q}}|^2}\right]
\end{equation}
where the division occurs entrywise.

In inverse problems such as non-Cartesian MRI and parallel MRI, where the measuremnt operator $\mbf A$ is not diagonalizable in the Fourier domain, we propose to solve (\ref{el}) efficiently using pre-conditioned conjugate gradient (CG) algorithm. The above simplifications of the derivative operator can be used to develop an efficient pre-conditioner in these cases. A few CG steps at each iteration are often sufficient for good convergence since the algorithm is initialized by the previous iterate.

\subsection{Efficient Evaluation of Shrinkage Weights}
We now focus on the efficient evaluation of $\mbf v_{\mbf q}(\mbf x); \,\forall \mbf q \in \cal N$ in (\ref{wt}). Note that $\mbf u_{\mbf q}(\mbf x)$ involves the comparison of the patches $P_{\mbf x}(\mbf f)$ and $P_{\mbf x+\mbf q}(\mbf f)$; since these quantities have to be computed for all spatial locations $\mbf x$ and different shifts $\mbf q$, the direct evaluation of (\ref{wt}) is computationally expensive. We propose to exploit the redundancies between $\mbf v_{\mbf q}(\mbf x)$ to considerably accelerate their computation. From (\ref{wt}), we have
\begin{eqnarray*}
\mbf u_{\mbf q}(\mbf x) &=& \nu\bkt{\|P_{\mbf x}(\mbf f) - P_{\mbf x+\mbf q}(\mbf f)\|}\\
&=& \nu\bkt{\sqrt{ \underbrace{\sum_{p \in \cal B}\|\mbf f(\mbf x-\mbf p) -  \mbf f(\mbf x-\mbf p+\mbf q)\|^{2}}_{(|{\mbf D}_{\mbf q}\mbf f|^{2}~*~\eta)(\mbf x)}}}.
\end{eqnarray*}
Here $\eta$ is a moving average filter with the size of the patch. The above equation implies that the computation of $\mbf u_{\mbf q}(\mbf x); \forall \mbf x$ can be obtained efficiently by simple pointwise operations and a computationally efficient filtering operation. Combining the above result with (\ref{wt}), we obtain
\begin{equation}\label{vq}
\mbf v_{\mbf q} = \underbrace{\left[\nu\bkt{\sqrt{|{\mbf D}_{\mbf q}\mbf f|^{2}~*~\eta}}\right]}_{\mbf u_{\mbf q}}~*~\eta
\end{equation}
We realize the convolutions $|{\mbf D}_{\mbf q}\mbf f|^{2}\ast\eta$ and $\mbf u_{\mbf q}\ast\eta$ using separable moving average filtering operations.
\subsection{Continuation Strategy to Improve Convergence}
The quality of the majorization in (\ref{new}) depends on the parameter $\beta$. It is known that high values of $\beta$ results in poor convergence. However, since we require the convex-hull approximation of the original penalty (see Section \ref{majsec}) for the majorization, the solution of the proposed scheme corresponds to that of the original problem only when $\beta \rightarrow \infty$. We hence use a continuation strategy to improve the convergence rate, where $\beta$ is initialized with a small value and is increased gradually to a high value. This approach is adopted by us \cite{Wendy,yangisbi11}, as well as other authors \cite{homotopy,wang2008new}. We also use continuation to truncate the metric penalties in which we start with a large threshold and gradually decrease it until it attains a small value; this means that we are not concerned with the distant patches that are high possible to be dissimilar. The saturated $\ell_{p}$ norm seems to give better results than non truncated one.

The pseudo-code of the algorithm is shown below.

\begin{pseudocode}[plain]{NonLocal Shrinkage}{\mathbf A,\mathbf b,\lambda}
\textbf{Input}: \mathbf b = \text{ k-space measurements }\\
\beta = \beta_{\rm init};\,T = T_{\rm init};\\
\WHILE i < \text{\# Outer Iterations}
\DO
\BEGIN
\WHILE j < \text{\# Inner Iterations}
\DO
\BEGIN
\text{Compute $\mbf v_{\mbf q};\forall \mbf q\in\mathcal N$ using \eqref{vq}}\\
\text{Compute $\mbf h_{\mbf q};\forall \mbf q\in\mathcal N$ using \eqref{hq}}\\
\text{Update $\mbf f$ according to \eqref{f}}\\

\END\\
\beta \GETS \beta*\beta_{\rm incfactor}\\
T \GETS T*T_{\rm decfactor}\\
\END\\
\RETURN{\mbf f}
\end{pseudocode}

We observe from the pseudo-code that the algorithm requires two moving-average filtering operations per $\mbf q$ value to evaluate (\ref{vq}). For a $3\times 3$ neighborhood, this translates to 16 moving filtering operations. In addition, the evaluation of $\mbf f$ according to (\ref{f}) requires one FFT and one IFFT.  We typically need 20 inner iterations and about 30--40 outer iterations for the best convergence and recovery.

The algorithm was implemented in MATLAB 2012 using Jacket \cite{jacket} on a Linux workstation machine with eight cores and a NVDIA Tesla graphical processing unit. In all the experiments, we initialize $\beta$ = 0.01 and update it by a factor of 2 in each outer iteration.

\section{Results}
We will now focus on the implementation of our scheme in the contexts of CS and denoising.
Some of the MR images used in these experiments are courtesy of American Radiology Services$\footnote[1]{ www3.americanradiology.com/pls/web1/wwimggal.vmg}$.
\subsection{Convergence Rate}
\begin{figure}[H]
\begin{center}
\subfigure[Cost vs Computation Time]{\includegraphics[width=4.25cm,height=3.5cm]{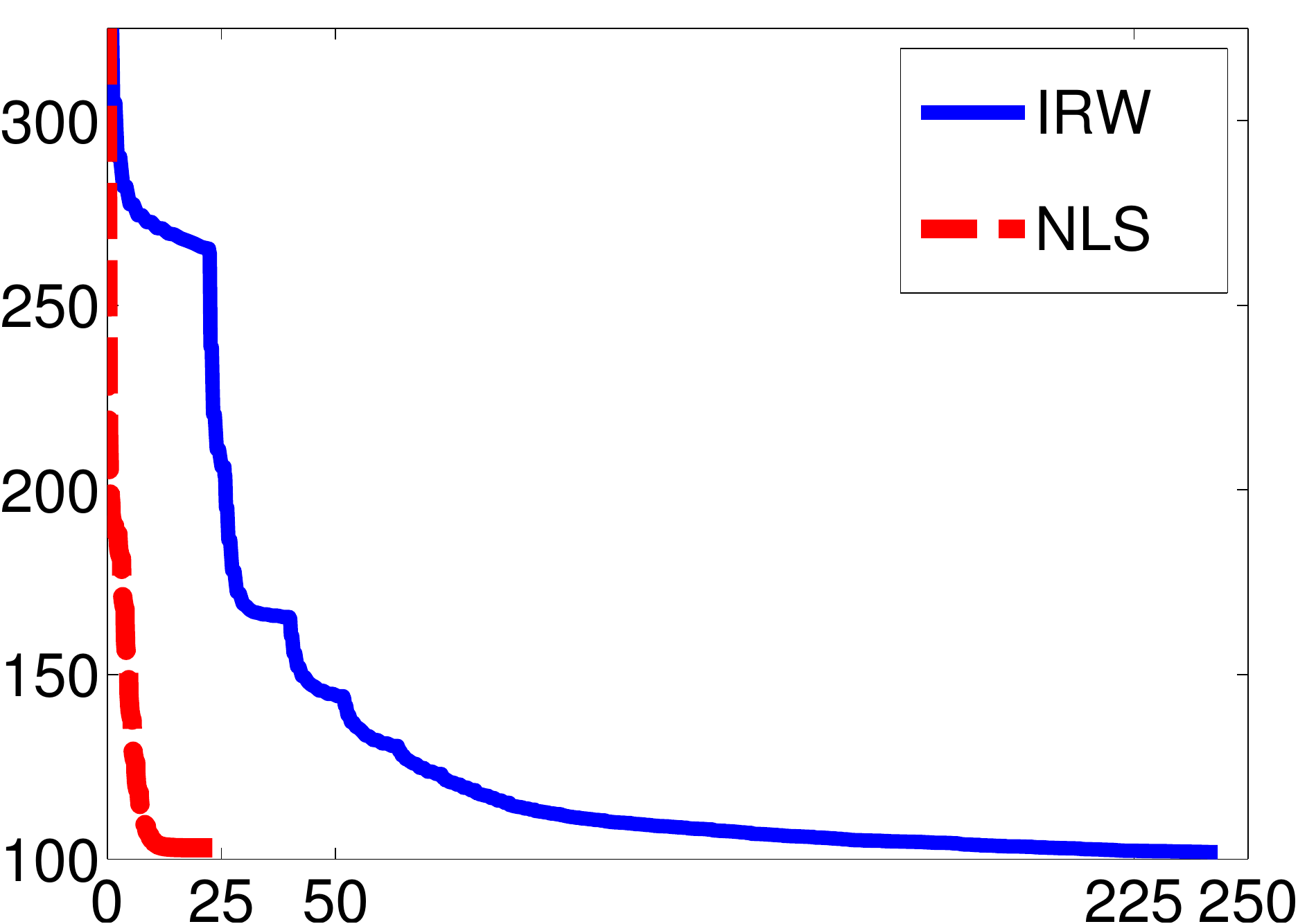}}
\subfigure[SNR vs Computation Time]{\includegraphics[width=4.25cm,height=3.5cm]{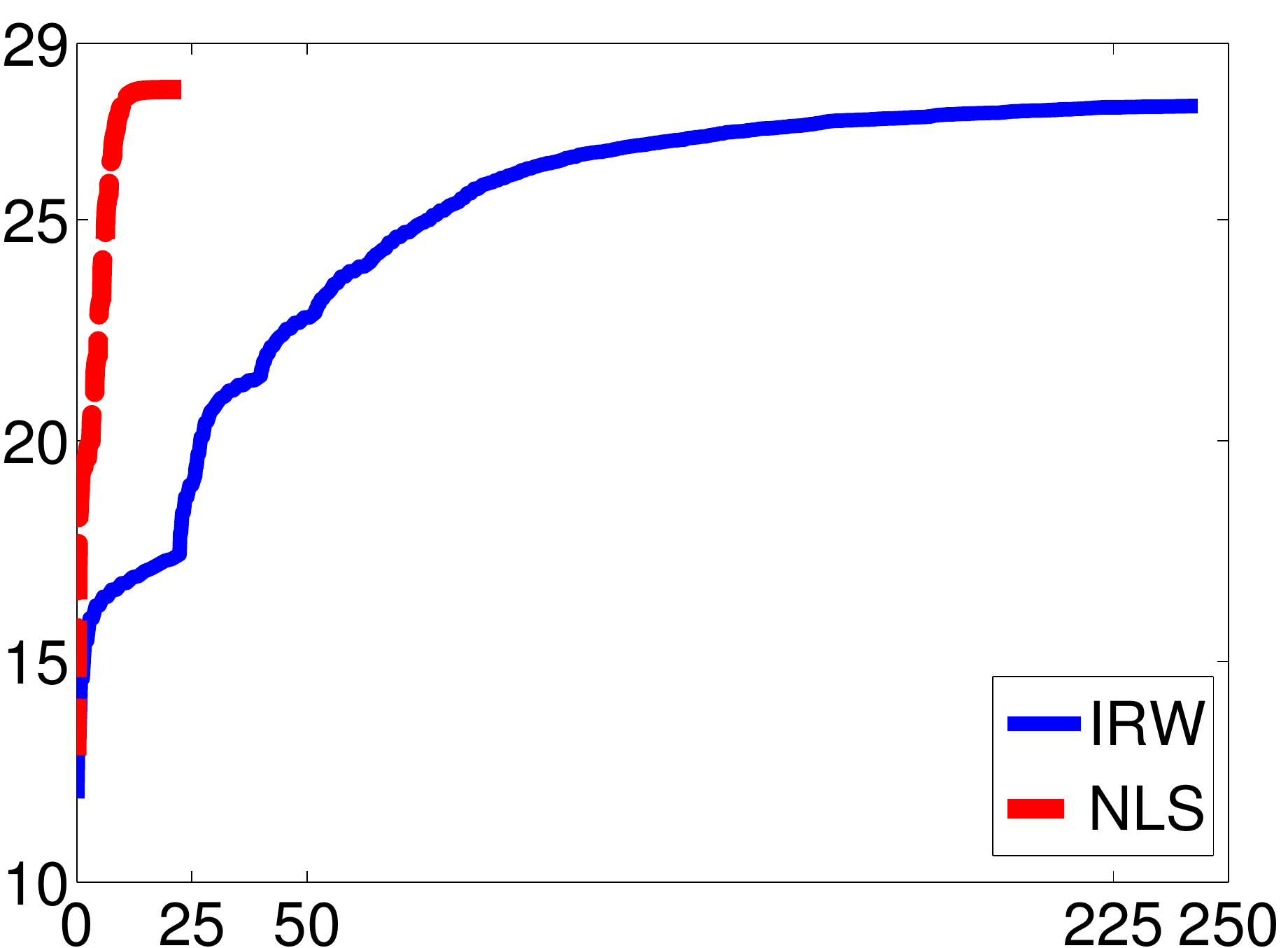}}
\end{center}
\caption{\small Comparison of the convergence of the iterative reweighted non-local algorithm and the proposed iterative non-local shrinkage algorithm. The plots indicate the evolution of the cost function specified by (\ref{cost}) and the signal to noise ratio (SNR) as a function of the computational time. We observe that the proposed scheme converges around ten times faster than the iterative reweighted algorithm.}
 \label{CONVSNR}
\end{figure}

We first compare the proposed scheme with our previous iterative reweighted non-local algorithm \cite{Wendy}. We consider the recovery of the 256$\times$256 MR brain image using a five fold under sampled random sampling pattern. The regularization parameters of both algorithms were set to  $\lambda =10^{-4}$; this parameter was chosen to obtain the best possible reconstruction by comparing with the original image. The convergence plots of the algorithm as a function of the CPU time are shown in Fig (\ref{CONVSNR}). We observe that both the algorithms converge to almost the same result. However, the non-local shrinkage algorithm converged around ten times faster than the iterative reweighted scheme. The reconstructions demonstrates the quality improvement offered by the proposed scheme for a specified computation time. One of the reasons for the faster convergence of the proposed algorithm can be attributed to the fast inversion of the quadratic sub-problems. The condition number of the quadratic subproblem in iterative reweighting \cite{Wendy} grows with iterations, resulting in slow convergence of the CG algorithms that were used to solve it.
\subsection{Impact of the Distance Metric}
The proposed scheme can be adapted to most non-local distance metrics by simply changing the shrinkage rule. The shrinkage rules for different non-local penalties are shown in Table I. We compare the different metrics in the context of recovering images from 5 fold acceleration using randomly under sampled data. The SNR of the reconstructions are shown in Table II. The regularization parameters of each of the algorithms are optimized to provide the best possible results. The first column corresponds to the convex $\ell_{1}$ differences between patches. The second and third columns correspond to alternating H1 and NLTV penalties \cite{Wendy}, respectively. These penalties depend on the parameter $\sigma$ corresponding to the width of the Gaussian weight function. This parameter is analogous to the threshold $T$ used in the saturating $\ell_{1}$ and $\ell_{p}$ penalties, shown in the last two columns. All of these parameters were optimized to ensure fair comparisons. We also used continuation schemes on these parameters to improve the convergence to global minima.

All of the penalties, except the $\ell_{1}$ distance function saturate with inter-patch distances. This explains the poor performance of the convex $\ell_{1}$ penalty compared to the non-convex counterparts. Unlike local total variation scheme, which only compares a particular pixel with a few other pixels, several pixel comparisons are involved in non-local regularization. Saturating priors are needed to avoid the averaging of dissimilar patches, which may result in blurring. Since the saturating $\ell_{p}$ metric
provides the best over all reconstructions, we use this prior for all the future comparisons in this paper.

\begin {table}[H]
\centering
\caption
{\small Impact of the distance metric on the reconstructions. We compare the reconstructions obtained using the non-local shrinkage algorithm using $\ell_{1}$, $H_{1}$, $NLTV$, thresholded $\ell_{1}$ and thresholded $\ell_{p}; p=0.5$ metrics. All the metrics, except the convex $\ell_{1}$ scheme are saturating priors. We observe that saturation is key to good performance of non-local algorithms. Among the different metrics, the thresholded $\ell_{p}$ penalty is observed to provide the best results in all the examples.}
\begin{tabular}{ | c || c | c | c | c | c | c |}
\hline \small \bf Image & \small \bf $\ell_{1}$ & \small $\mbf H_1$ &  \small\bf NLTV &\small\bf $\ell_{1}$-T & \small\bf $\ell_{p}$-T\\ \hline
 \hline
\small Brain1  & 22.12 &27.15 & 28.06 &27.53 & \textbf{28.41}\\
\small Brain2  & 22.48 &26.12  & 27.15&26.35& \textbf{27.93}\\
\small ankle  & 22.23 &23.43  &24.56 & 23.81& \textbf{24.67}\\
 \hline
 \end{tabular}
\end {table}

\subsection{Comparisons With State-of-the-Art Algorithms}
We compare the proposed scheme with classical local total variation algorithm and the dictionary learning MRI (DLMRI) scheme \cite{DLMRI}. The DLMRI method is also a patch based regularization scheme, where a dictionary is learned from the patches in the image. This scheme was reported \cite{DLMRI} to provide considerably better reconstructions than the sparse recovery algorithm combining wavelet and TV regularization \cite{LDP}. We relied on the MATLAB implementation of DLMRI$\footnote[2]{The DLMRI code is available on the author's website www.ifp.illinois.edu/~yoram/DLMRI-Lab/DLMRI.html}$. A key difference with the results reported in \cite{DLMRI} is that we used the complex version of the code distributed by the authors. This was required to make the comparisons fair to TV and our non-local implementations as both of them do not use this constraint. Note that this assumption is often not satisfied in routine MRI exams.

\begin{figure*}
\vspace{-7mm}
\centering
\subfigure[Original]{\includegraphics[width=0.23\textwidth,height=0.23\textwidth,trim=1cm 1cm 1cm 1cm,clip]{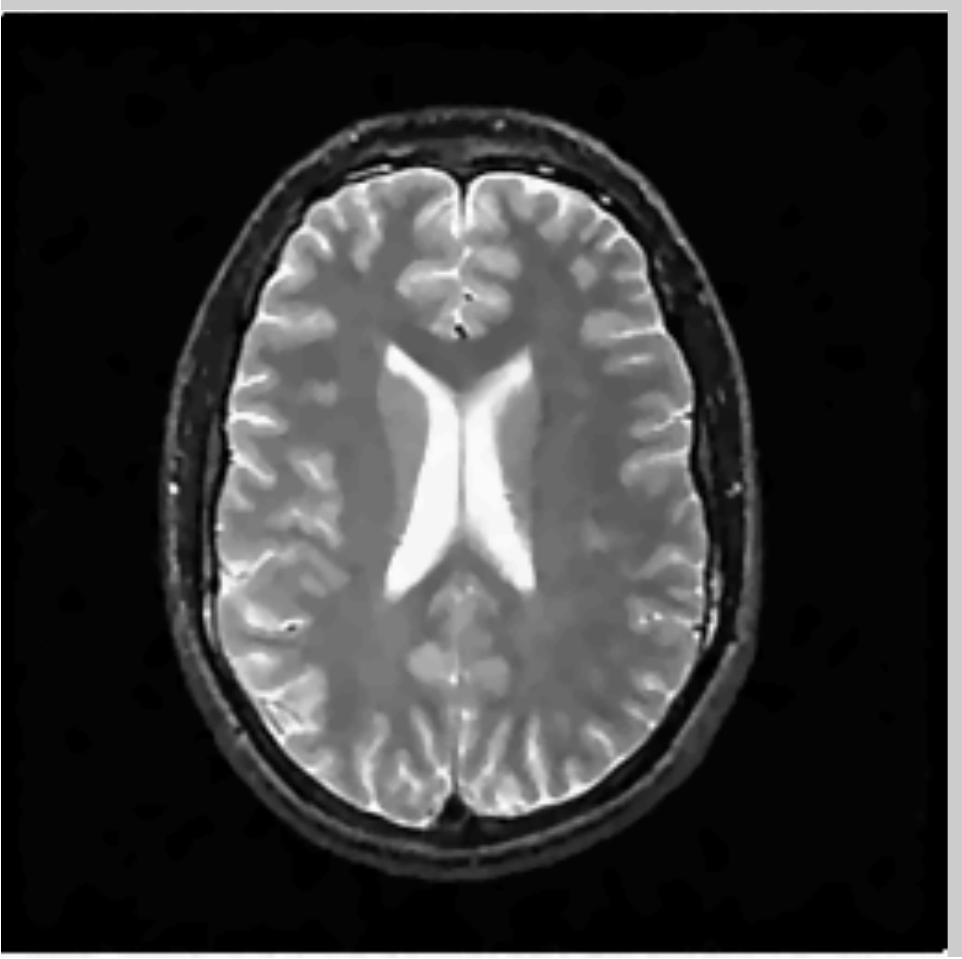}}
\subfigure[DLMRI, SNR=20.46]{\includegraphics[width=0.23\textwidth,height=0.23\textwidth,trim=1cm 1cm 1cm 1cm,clip]{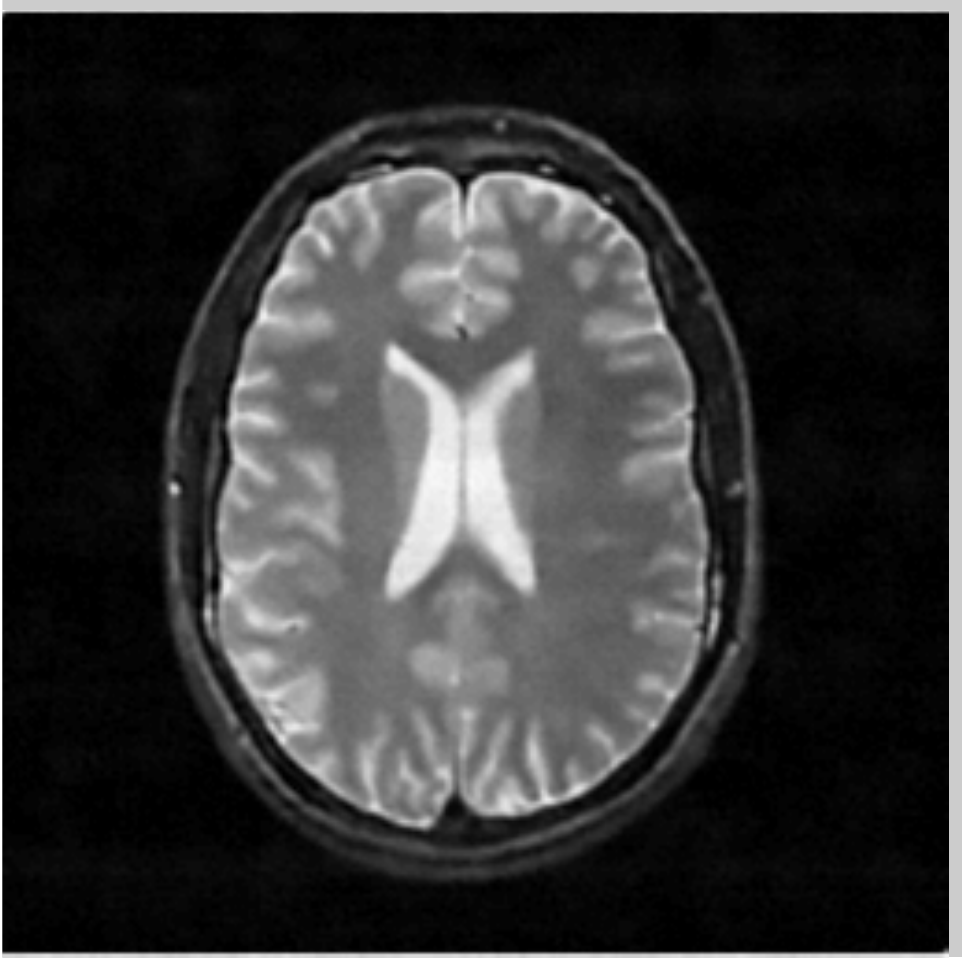}}
\subfigure[TV, SNR=22.80]{\includegraphics[width=0.23\textwidth,height=0.23\textwidth,trim=1cm 1cm 1cm 1cm,clip]{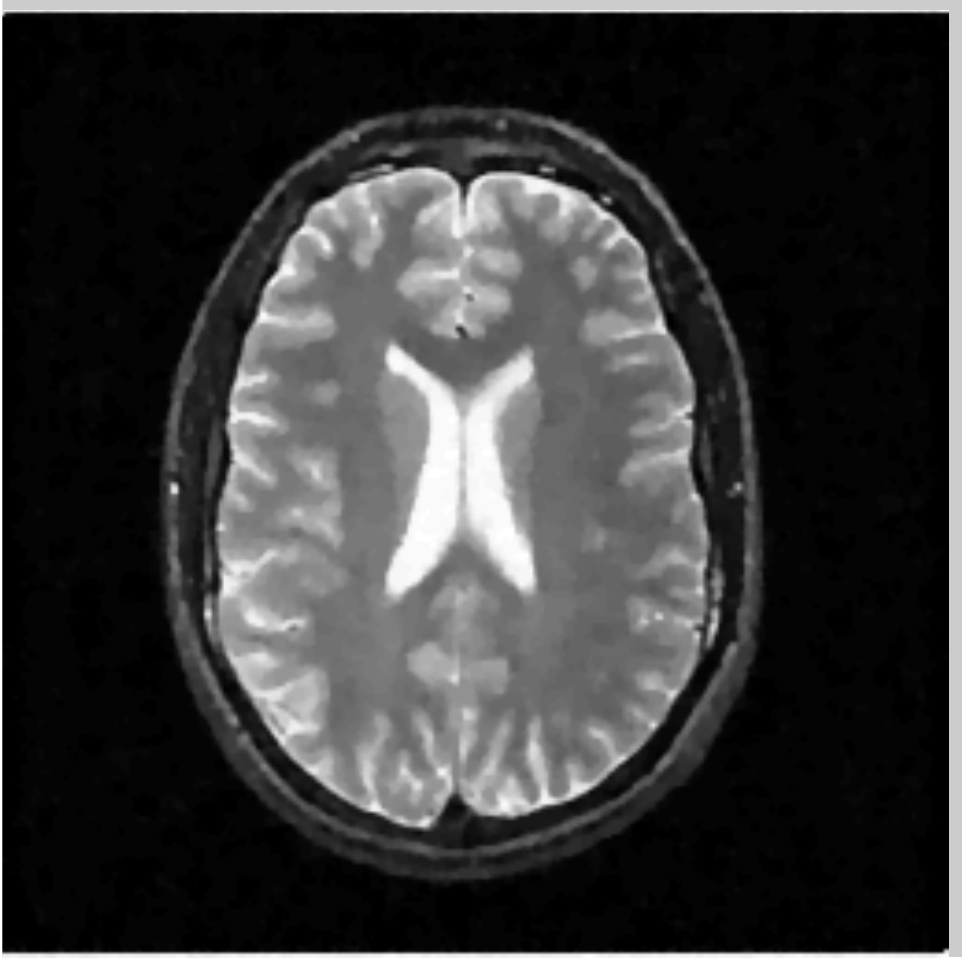}}
\subfigure[NLS, SNR=28.41]{\includegraphics[width=0.23\textwidth,height=0.23\textwidth,trim=1cm 1cm 1cm 1cm,clip]{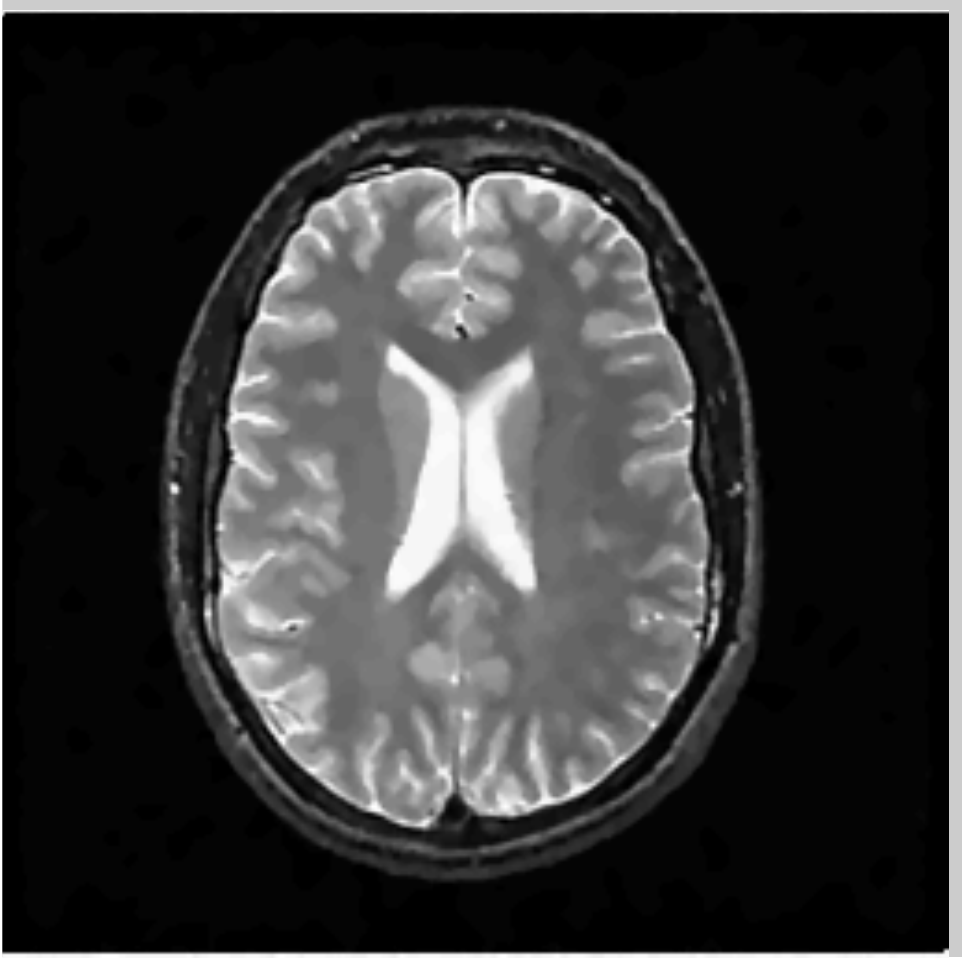}}
\\
\subfigure[Sampling pattern]{\includegraphics[width=0.23\textwidth,height=0.23\textwidth]{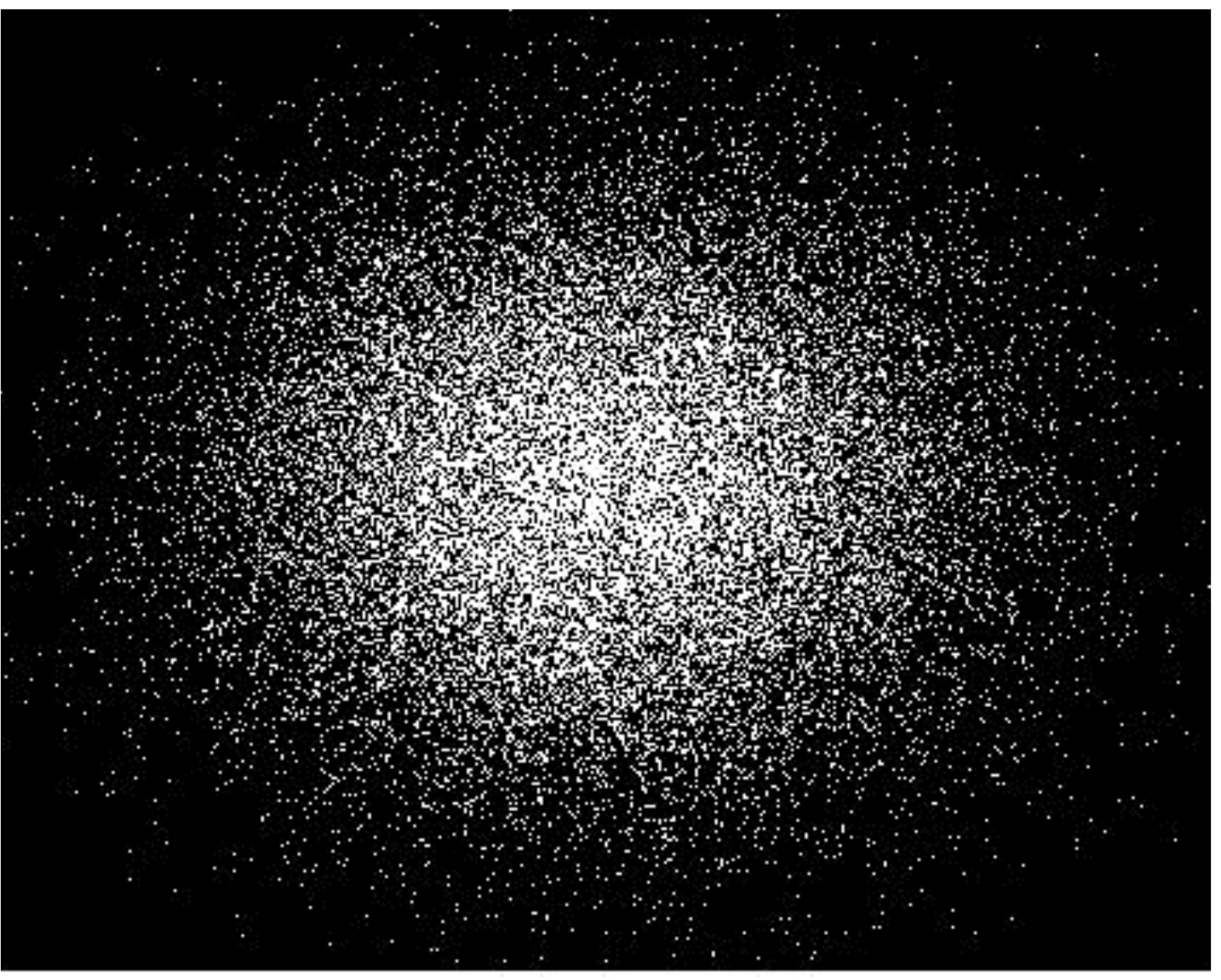}}
\subfigure[DLMRI error]{\includegraphics[width=0.23\textwidth,height=0.23\textwidth,trim=1cm 1cm 1cm 1cm,clip]{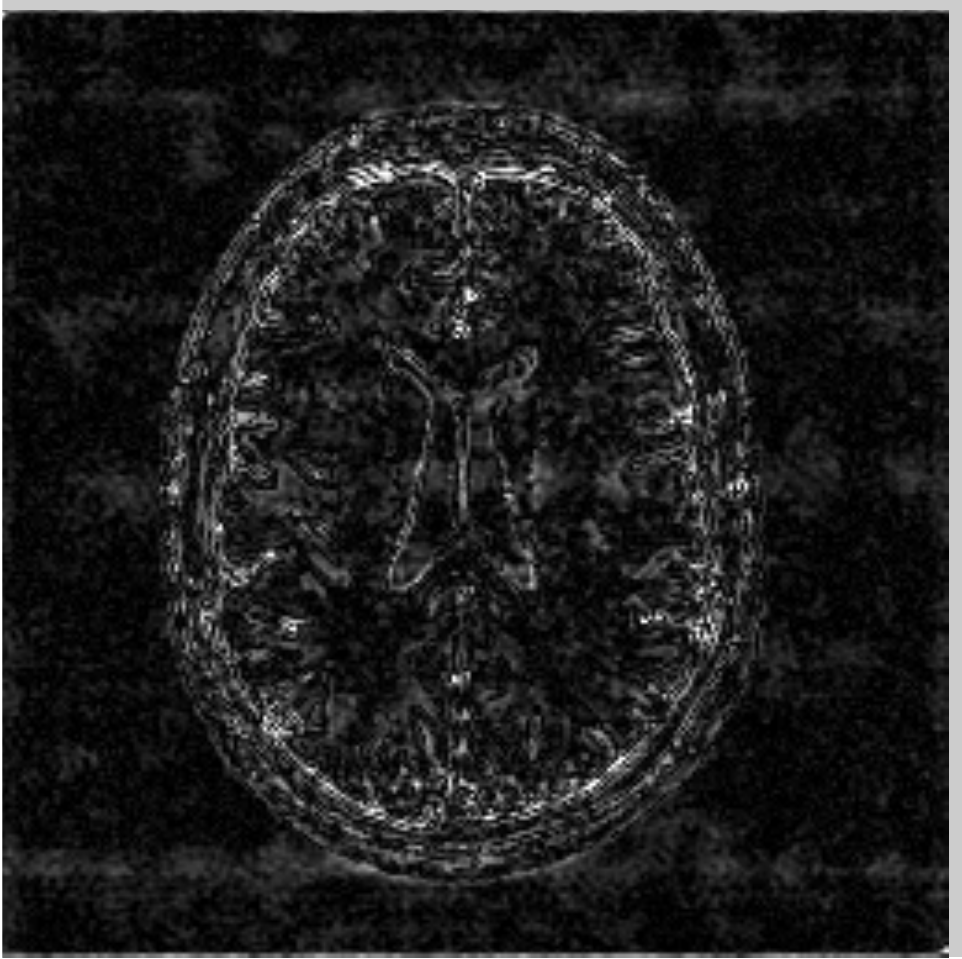}}
\subfigure[TV error]{\includegraphics[width=0.23\textwidth,height=0.23\textwidth,trim=1cm 1cm 1cm 1cm,clip]{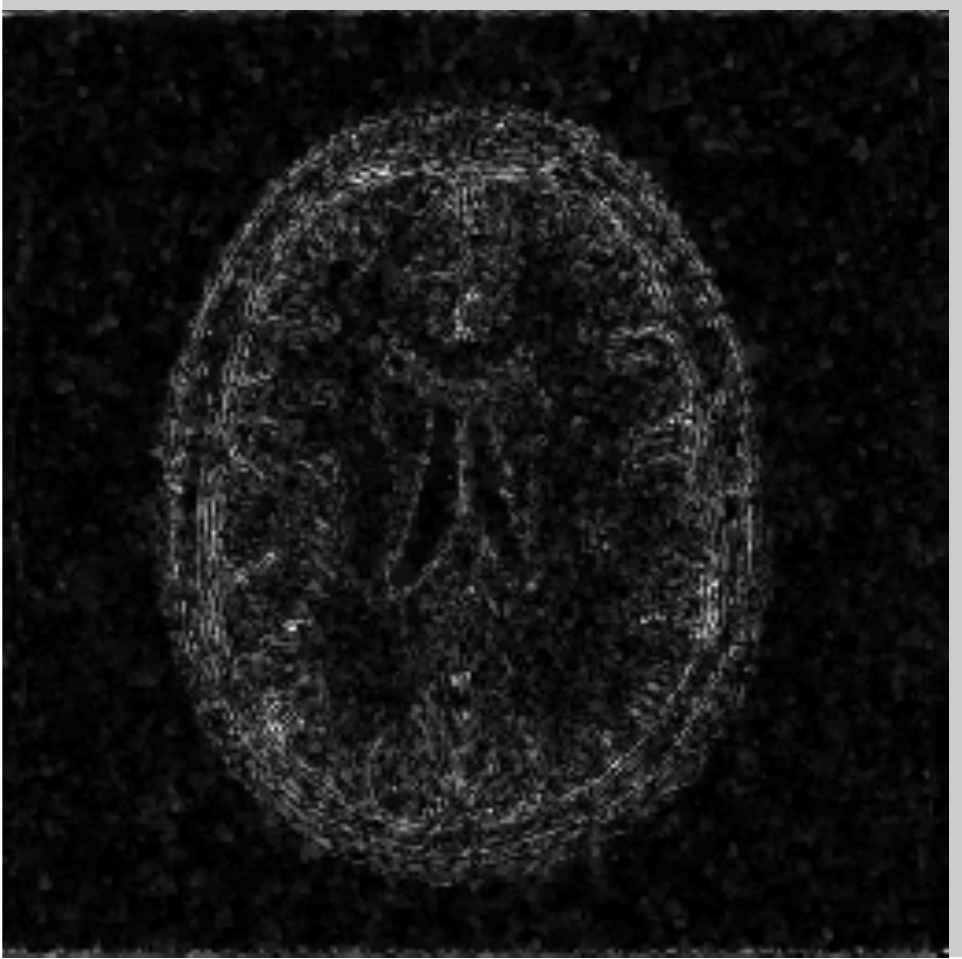}}
\subfigure[NLS error]{\includegraphics[width=0.23\textwidth,height=0.23\textwidth,trim=1cm 1cm 1cm 1cm,clip]{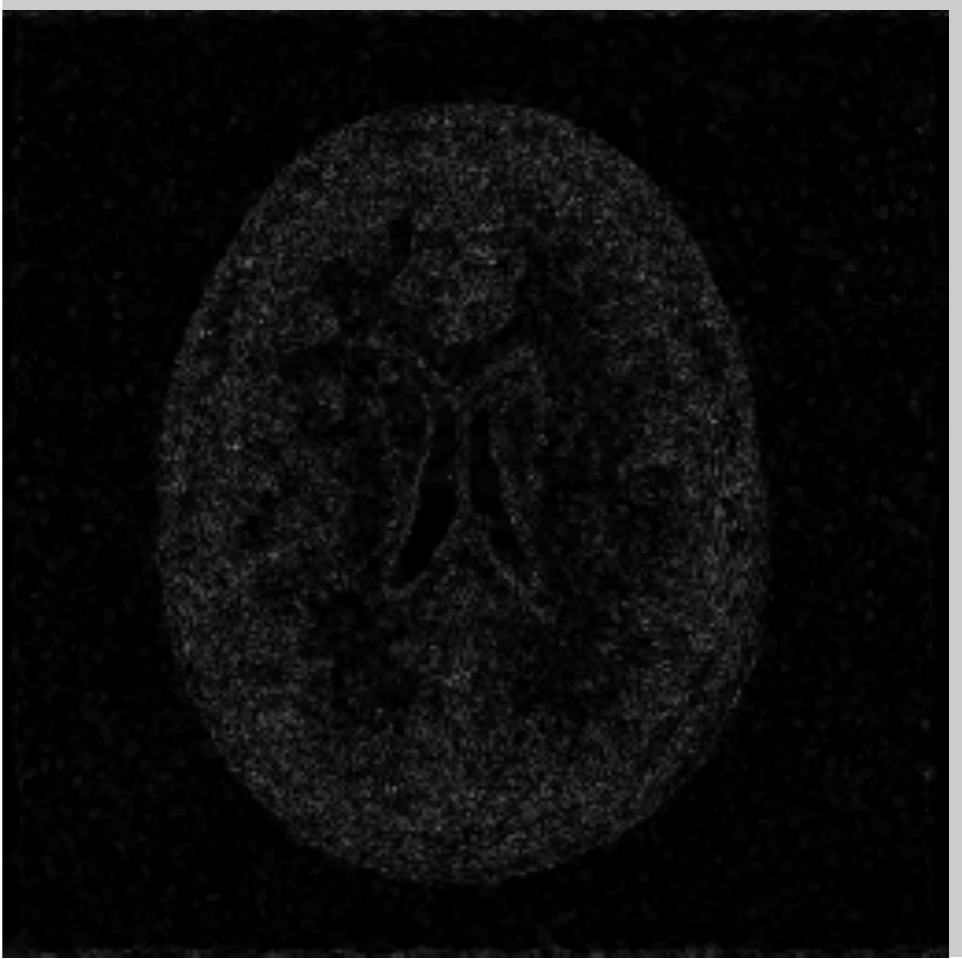}}

\caption{\small Comparison of the algorithms in the absence of noise. We consider the recovery of a 256$\times$256 MRI brain image from 20\% of its Fourier samples, acquired using a random sampling pattern shown in (e) using non-local shrinkage scheme (NLS), DLMRI and local TV (TV). The reconstructions are shown in (b)-(d). The corresponding error images, scaled by a factor of 5 for better visualization, are shown in the bottom row. The reconstructions show that the NLS scheme is capable of better preserving the edges and details, resulting in less blurred reconstructions.}
 \label{nonoise}
\end{figure*}

The comparison of the methods in the context of random sampling with 20\% of the samples retained in the absence of noise is shown in Fig. \ref{nonoise}. The regularization parameters of all the algorithms have been optimized to yield the best signal to noise ratio (SNR). The SNR and the peak SNR (PSNR) that are used for the comparisons in this paper are computed as
\begin{align*}
{\rm SNR} &= 20\log_{10} \bkt{\frac{||\mathbf{\Gamma}_{\rm orig}||_{F}}{||\mathbf{\Gamma}_{\rm rec}-\mathbf{\Gamma}_{\rm orig}||_{F}}}\\
{\rm PSNR} &= 20\log_{10} \bkt{\frac{{MAX}\cdot\sqrt{N_{x}\cdot N_{y}}}{||\mathbf{\Gamma}_{\rm rec}-\mathbf{\Gamma}_{\rm orig}||_{F}}}
\end{align*}
where $N_x$ and $N_y$ are the image dimensions, and $MAX$ is the maxiumum allowed pixel intensity.

We observe that the proposed non-local algorithm provides better preservation of edge details and minimize patchy artifacts as seen in TV reconstructions. The quantitative comparisons of different methods on more MR images in the absence of noise using 5 fold random sampling operator are reported in Table \ref{nonoisetable}. We observe that the NLS scheme provides a consistent 2-4 dB improvement over the other methods in most cases.

\begin{table}
\centering
\caption
{\small Quantitative comparison of the proposed iterative non-local shrinkage (NLS) algorithm using the saturating $\ell_{p};p=0.5$ penalty with dictionary learning MRI (DLMRI) \cite{DLMRI} and local total variation regularization schemes in the absence of noise. We considered random sampling. The SNR and PSNR metrics of the reconstructed images are shown in the table.}
\label{nonoisetable}
\begin{tabular}{c| c c | c c| c c }
\toprule
\bf Image &  \multicolumn{2}{c|}{\bf DLMRI} & \multicolumn{2}{c|}{\bf TV} & \multicolumn{2}{c}{\bf NLS}\\
\midrule
\hline
   & \small SNR   & \small PSNR    & \small SNR   & \small PSNR & \small SNR  & \small PSNR\\
\hline
\small Brain1   &  20.46 & 29.73 &22.80 & 32.87 & \textbf{28.41} & \textbf{39.14}\\
\small Brain2   &  21.40 & 31.77 & 23.61 & 34.66 &  \textbf{27.93} & \textbf{39.54}\\
\small Brain3   &  23.24 & 36.96 & 26.70 & 41.20 & \textbf{29.13} & \textbf{44.15} \\
\small Ankle   &  20.10 & 32.15 & 22.22 & 34.83 &  \textbf{24.67} & \textbf{38.00} \\
\small Spine   &  26.53 & 36.76 & 30.39 & 41.52 & \textbf{33.11} & \textbf{44.73} \\
\small Willis'  & 21.33 & 33.13 & 21.42 & 33.95  &\textbf{23.81} & \textbf{36.91} \\
\bottomrule
\end{tabular}
\end{table}

\subsection{Performance With Noise}
\begin{figure}
\begin{center}
\subfigure[PSNR vs Accelerations]{\includegraphics [width=4.25cm,height=3.5cm]{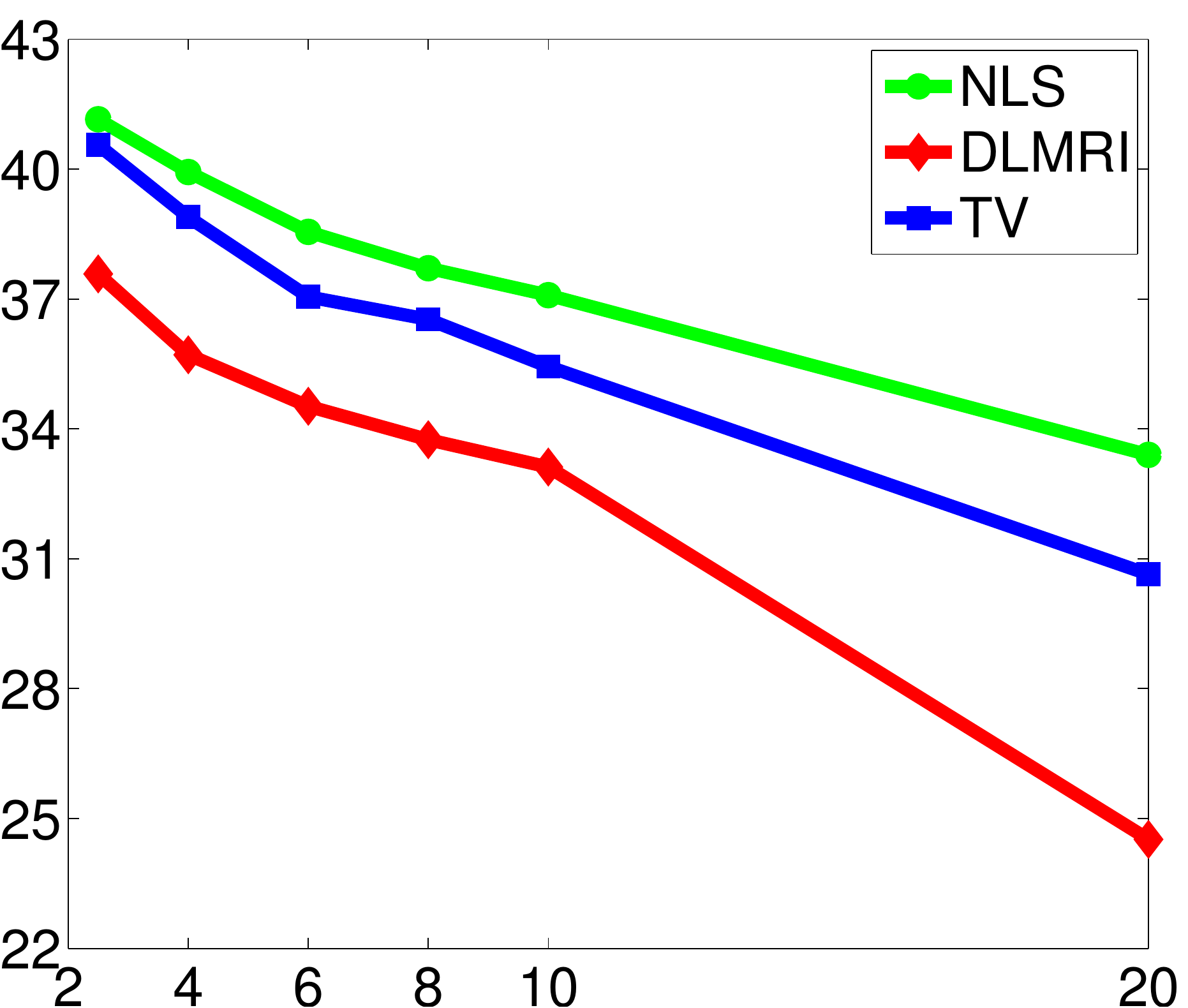}}
\subfigure[SNR vs Accelerations]{\includegraphics [width=4.25cm,height=3.5cm]{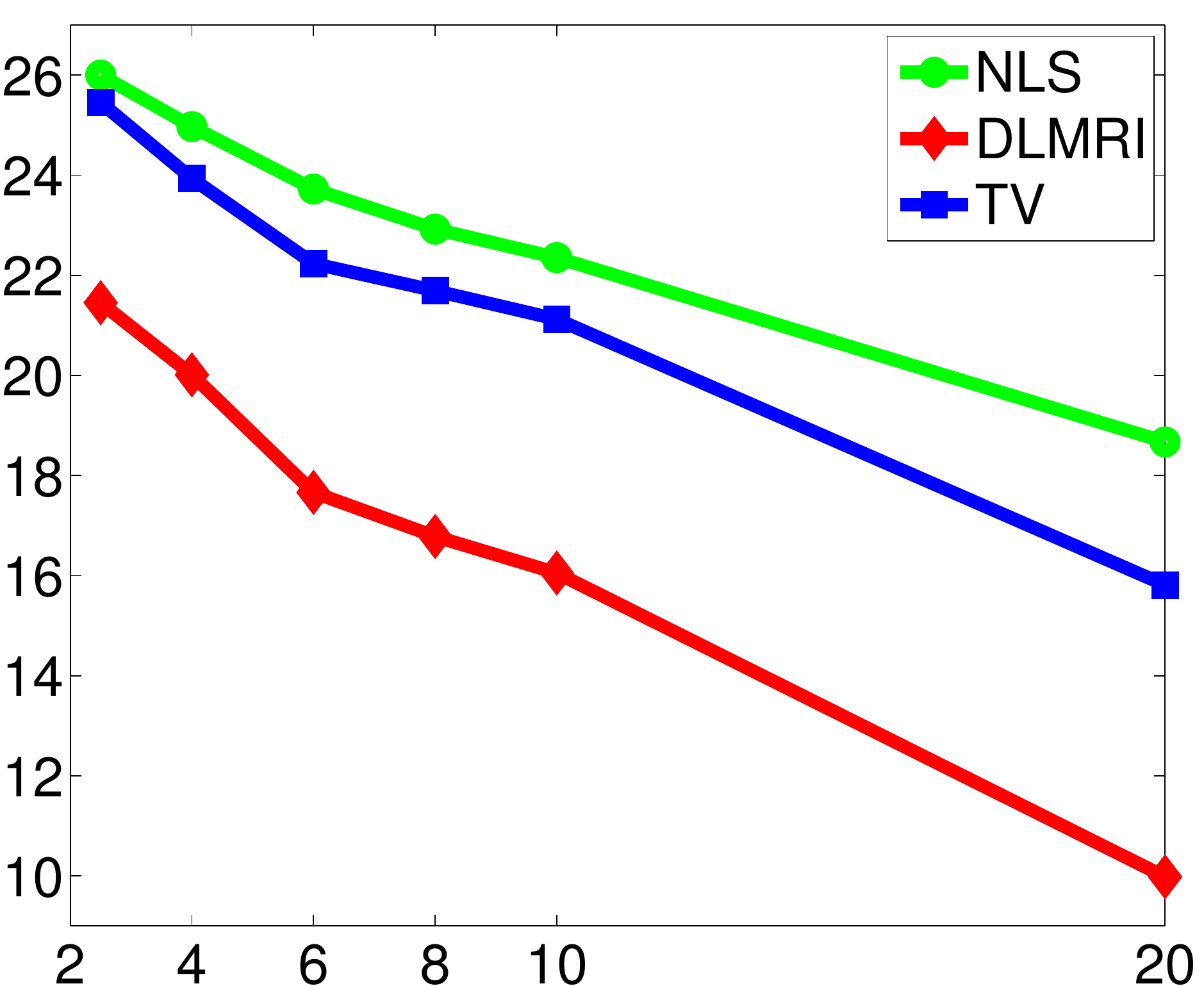}}
\end{center}
\caption{\small SNR and PSNR vs Acceleration. We used a $512\times 512$ MRI brain image, sampled using a random sampling operator at different acceleration factors (R=2.5,4,6,8,10 and 20). The measurements were contaminated with complex white Gaussian noise of $\sigma=10.2$. The SNR of the reconstructions obtained using the three algorithms are plotted. These results show that the NLS scheme is capable of providing better reconstructions at a range of accelerations. Specifically, it provides a consistent improvement of 0.5-9 dB in SNR/PSNR over the other algorithms.}
\label{psnrplots}
\end{figure}

We study the performance of the proposed algorithms as a function of acceleration in the presence of noise in Fig. \ref{psnrplots}.  We used a $512\times 512$ MRI brain image, sampled using a random sampling operator at different acceleration factors (R = 2.5, 4, 6, 8, 10 and $20$). The measurements were contaminated with complex white Gaussian noise of $\sigma=10.2$. The PSNR and SNR as a function of accelerations of this experiment are plotted in Fig. \ref{psnrplots}, where we compare our method against DLMRI and TV. We observe that the proposed scheme provides a consistent improvement in the presence of noise.

\begin{figure*}
\vspace{-7mm}
  \centering
\subfigure[Original]{\includegraphics[width=0.23\textwidth,height=0.23\textwidth]{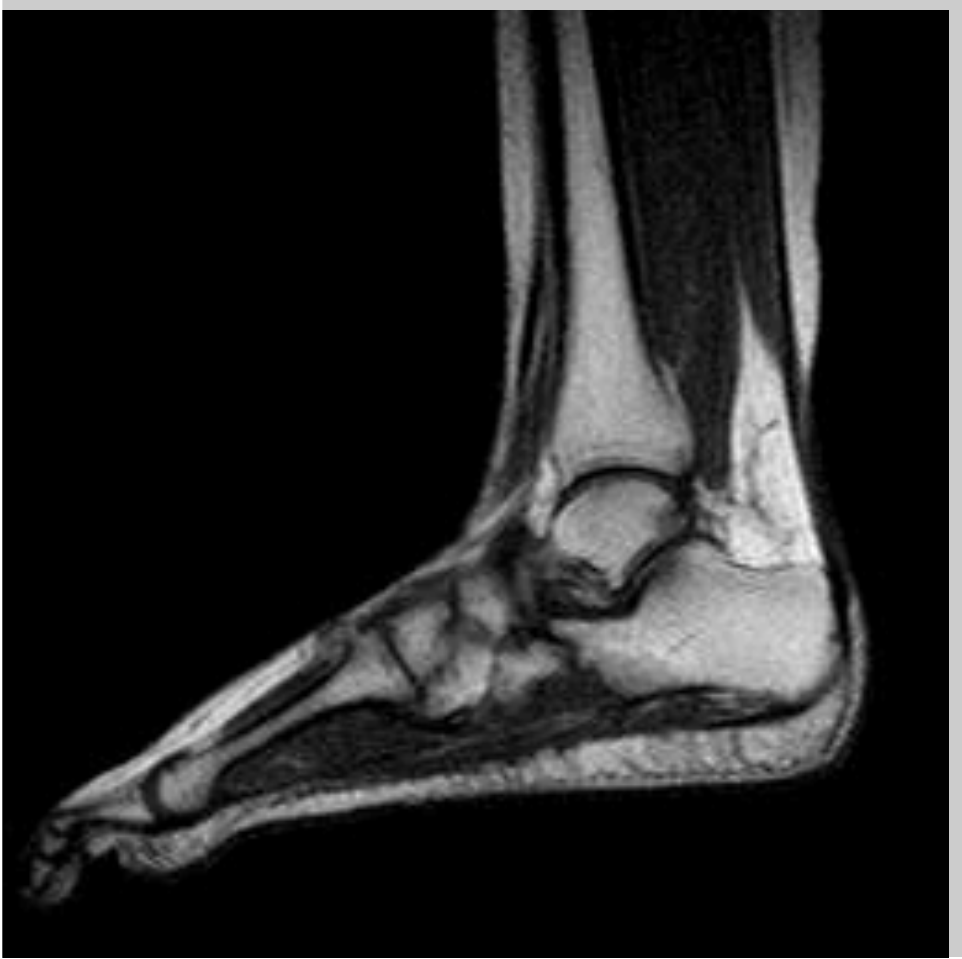}}
\subfigure[DLMRI, SNR=12.96]{\includegraphics[width=0.23\textwidth,height=0.23\textwidth]{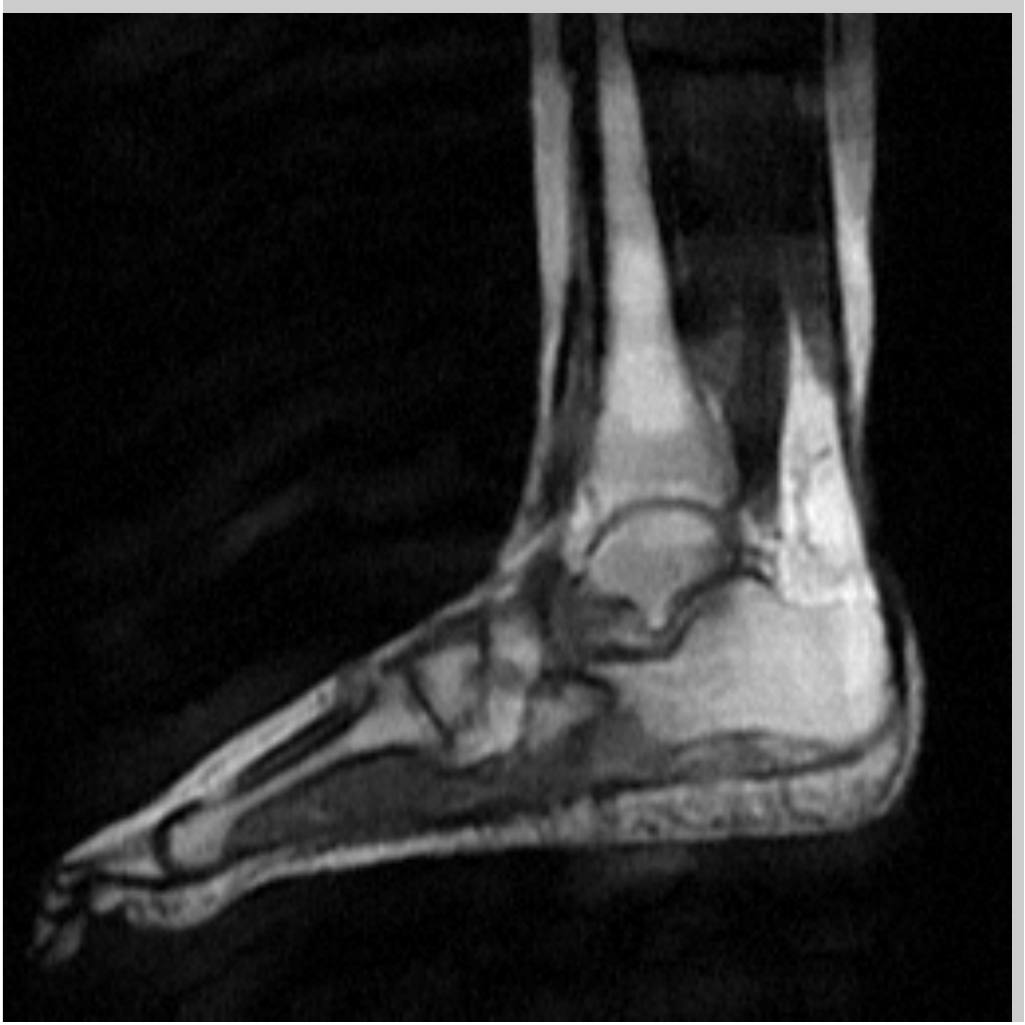}}
\subfigure[TV, SNR=15.02]{\includegraphics[width=0.23\textwidth,height=0.23\textwidth]{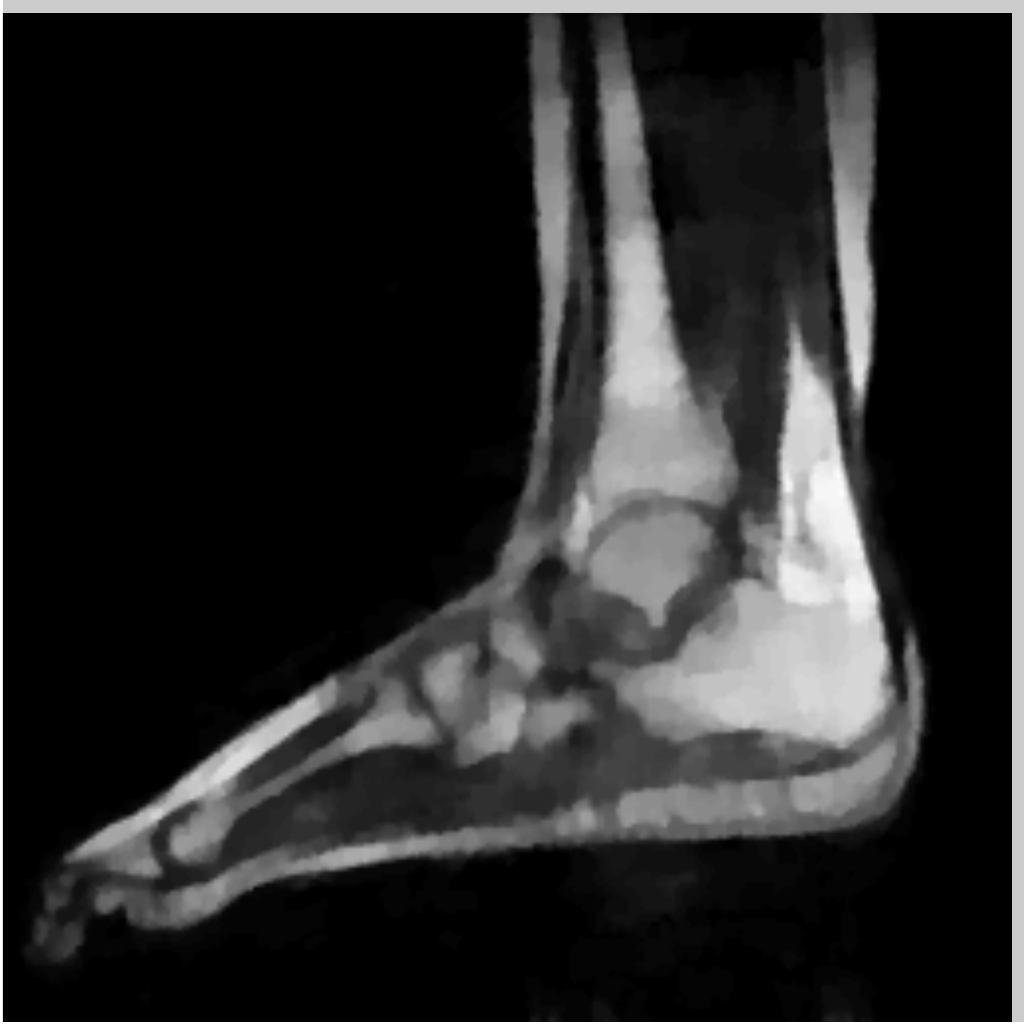}}
\subfigure[NLS, SNR=18.52]{\includegraphics[width=0.23\textwidth,height=0.23\textwidth]{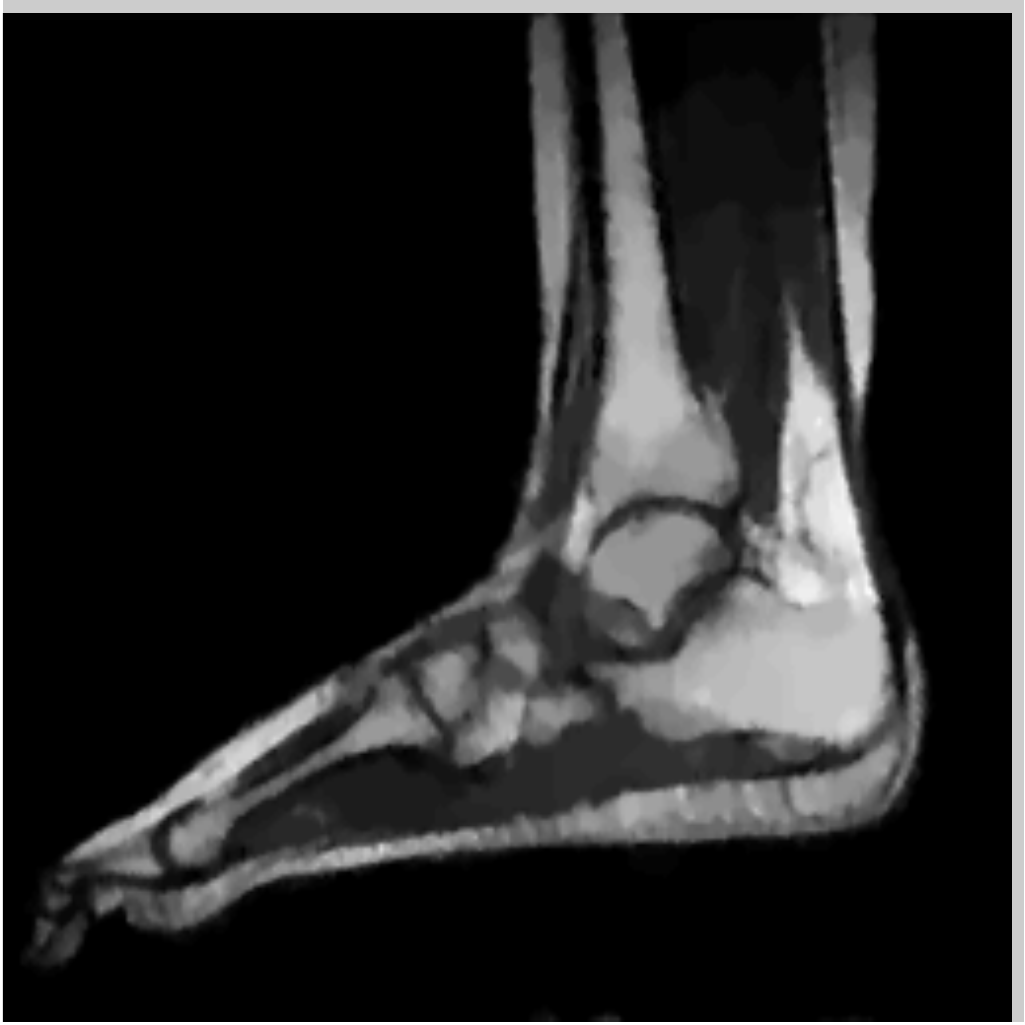}}\\
\subfigure[Sampling pattern]{\includegraphics[width=0.23\textwidth,height=0.23\textwidth]{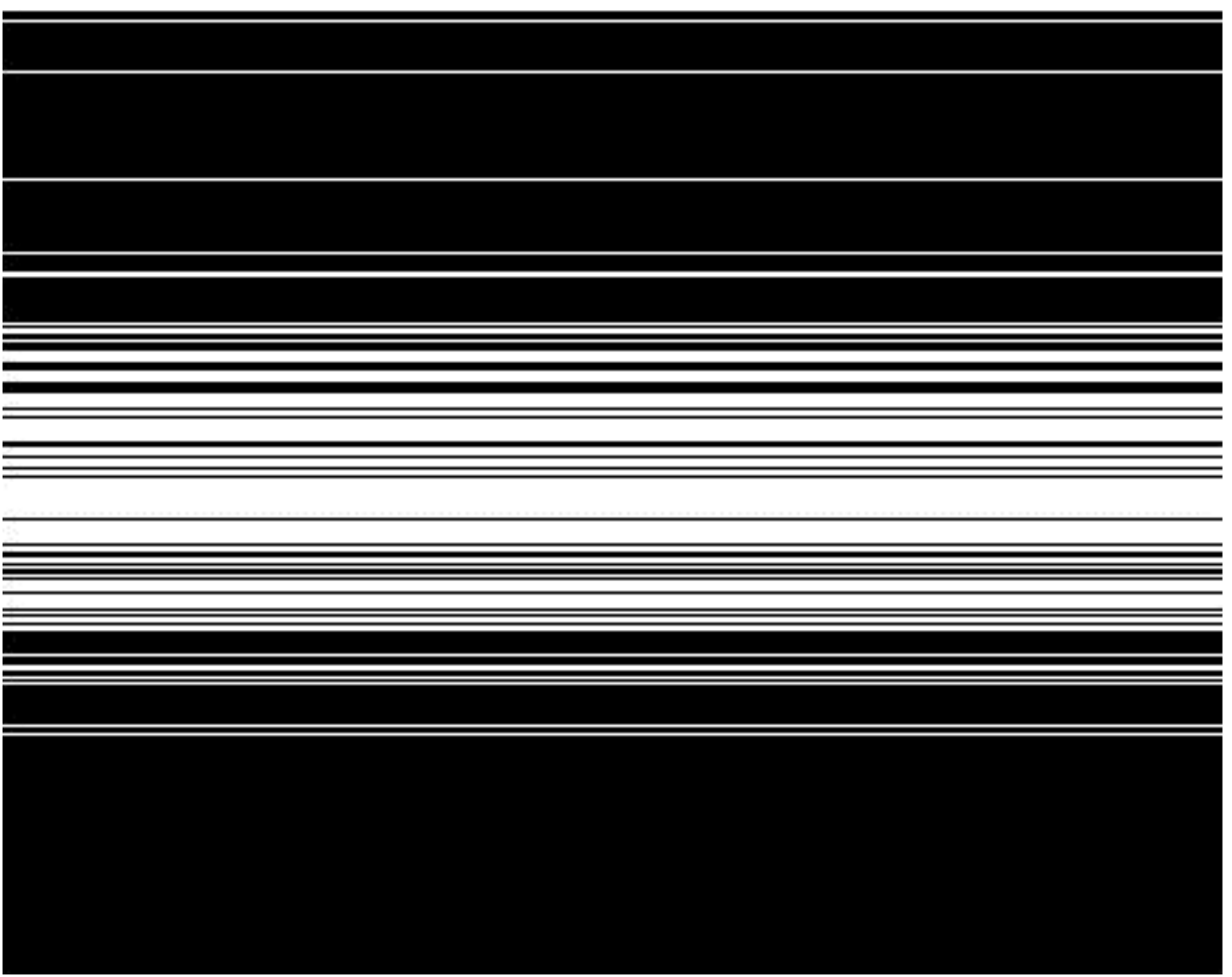}}
\subfigure[DLMRI error]{\includegraphics[width=0.23\textwidth,height=0.23\textwidth]{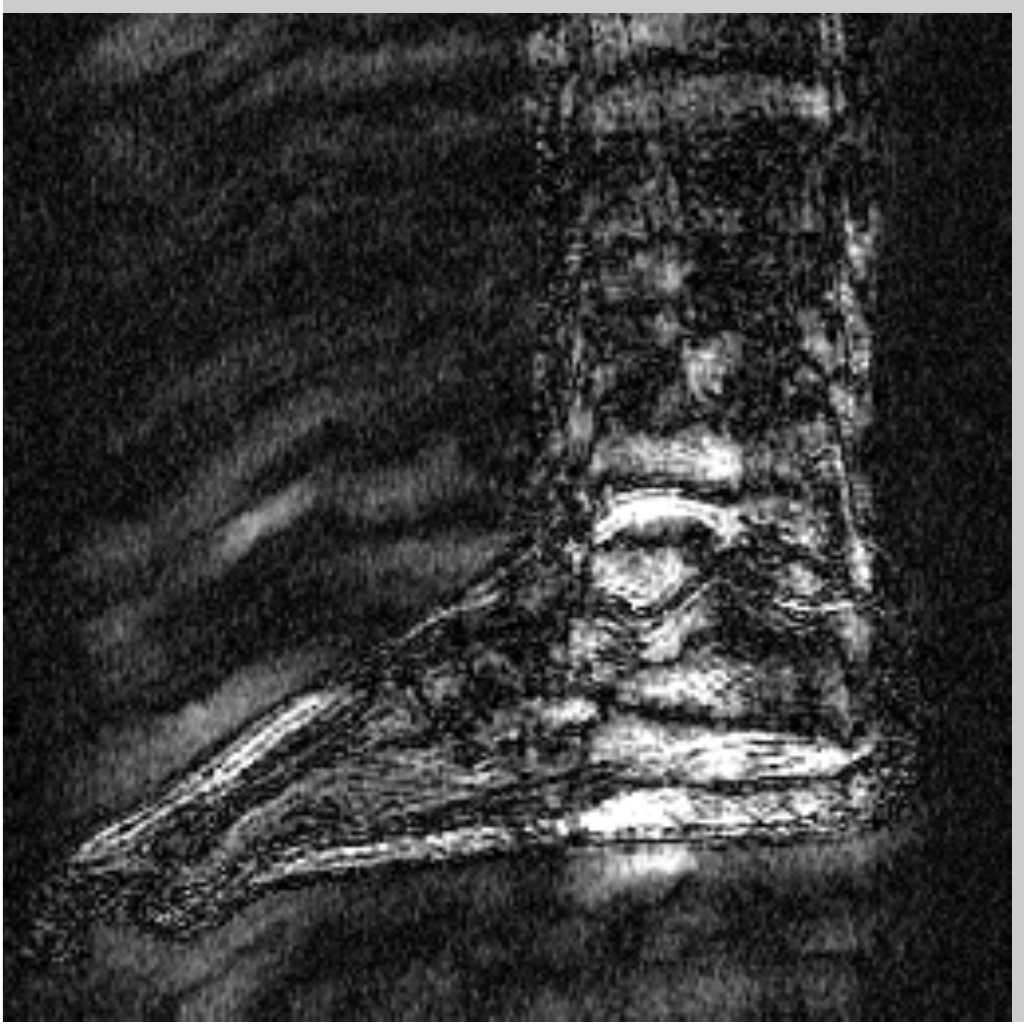}}
\subfigure[TV error]{\includegraphics[width=0.23\textwidth,height=0.23\textwidth]{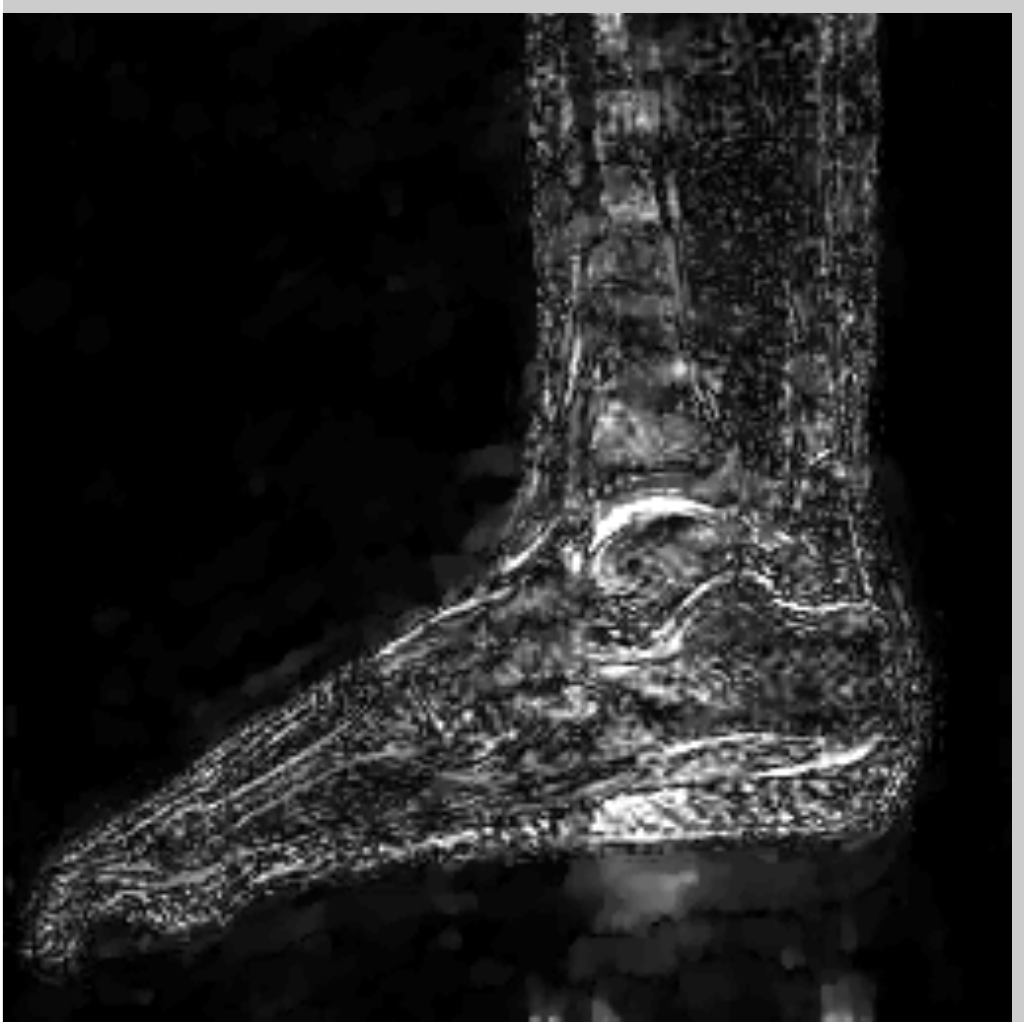}}
\subfigure[NLS error]{\includegraphics[width=0.23\textwidth,height=0.23\textwidth]{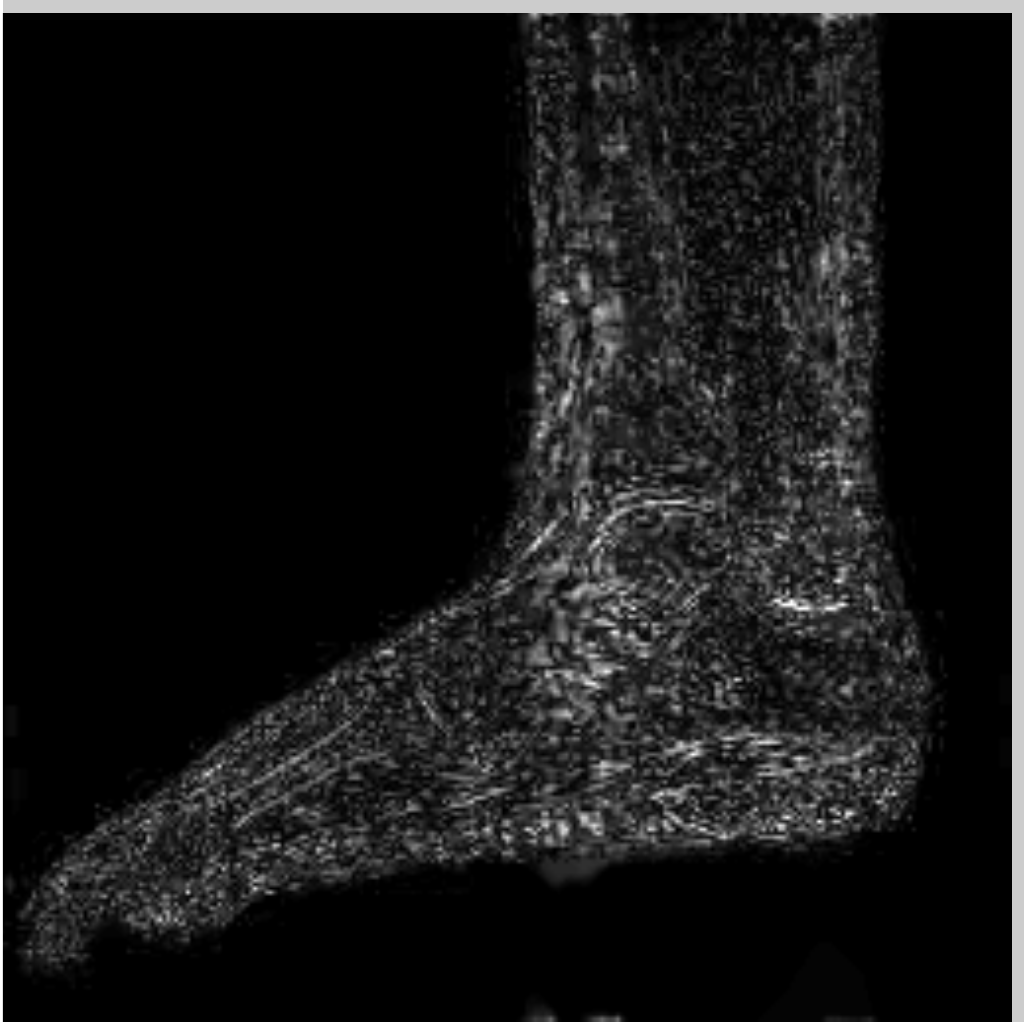}}

\caption{\small Comparison of the algorithms in the presence of noise. We consider the recovery of a $256\times 256$ MRI ankle image from its Cartesian Fourier sampling pattern (shown in (e)), contaminated by zero mean complex Gaussian noise with standard deviation $\sigma=10$. The top row shows the original and reconstructed images, while the error images scale by a factor of five are shown in the bottom row. This is a challenging case due to the high 1-D undersampling factors and noise. We observe that the NLS scheme provides the best reconstructions with minimal alias artifacts.}
\label{ankle}
\end{figure*}

The reconstructions of an ankle image from its 4 fold Cartesian undersampled Fourier data, corrupted with zero mean complex Gaussian noise with a standard deviation $\sigma=10$, are shown in Fig. \ref{ankle}. This is a really challenging case since the 1-D downsampling pattern is considerably less efficient than the 2-D random pattern used in the previous experiment. We observe that the non-local algorithm provides better reconstructions than the other schemes. Specifically, the TV scheme results in patchy artifacts. The DLMRI scheme results in blurring and loss of details close to the heel. The details are relatively better preserved close to the finger since there are no structures above or below it that aliases to it. By contrast to the classical algorithms, the degradation in performance of the non-local algorithm is comparatively small. The quantitative comparisons of the algorithms on this setting using different images are shown in the top section of Table IV.

\begin{figure*}
\vspace{-7mm}
\centering
\subfigure[Original]{\includegraphics[width=0.23\textwidth,height=0.23\textwidth,trim=0.5cm 0.5cm 0.5cm 0.5cm,clip]{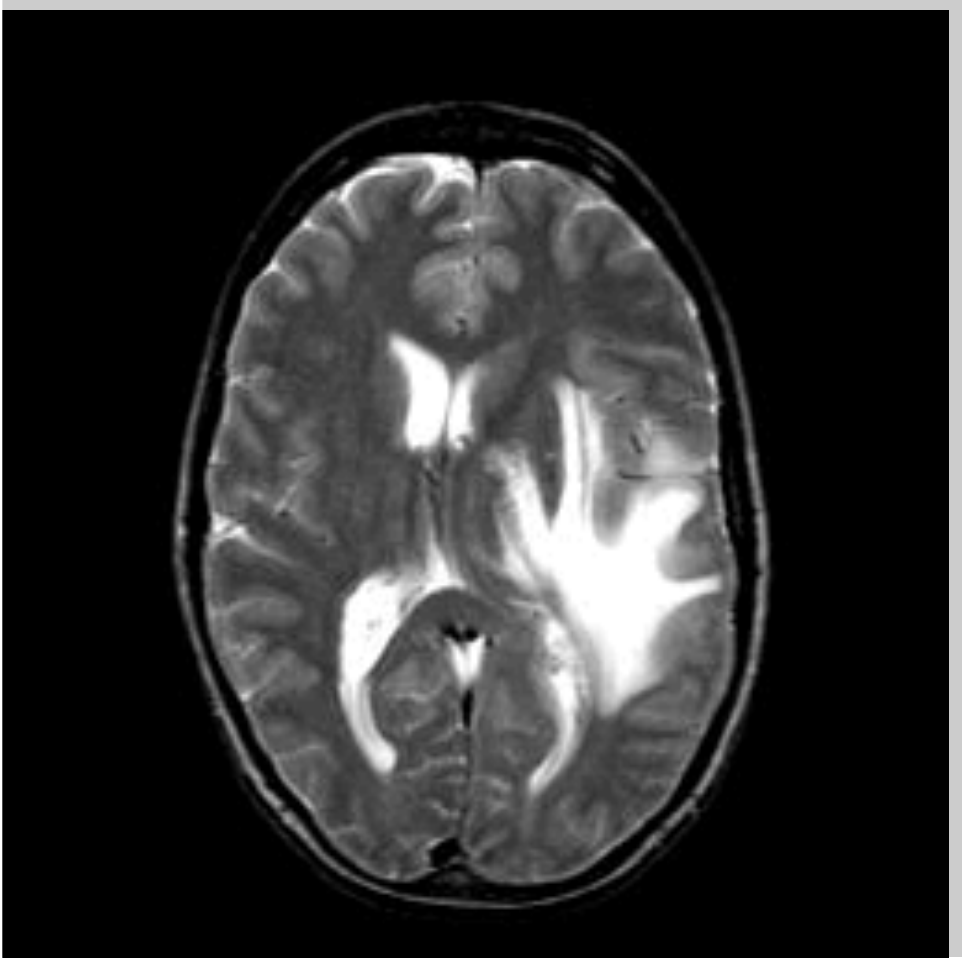}}
\subfigure[DLMRI,SNR=17.46]{\includegraphics[width=0.23\textwidth,height=0.23\textwidth,trim=0.5cm 0.5cm 0.5cm 0.5cm,clip]{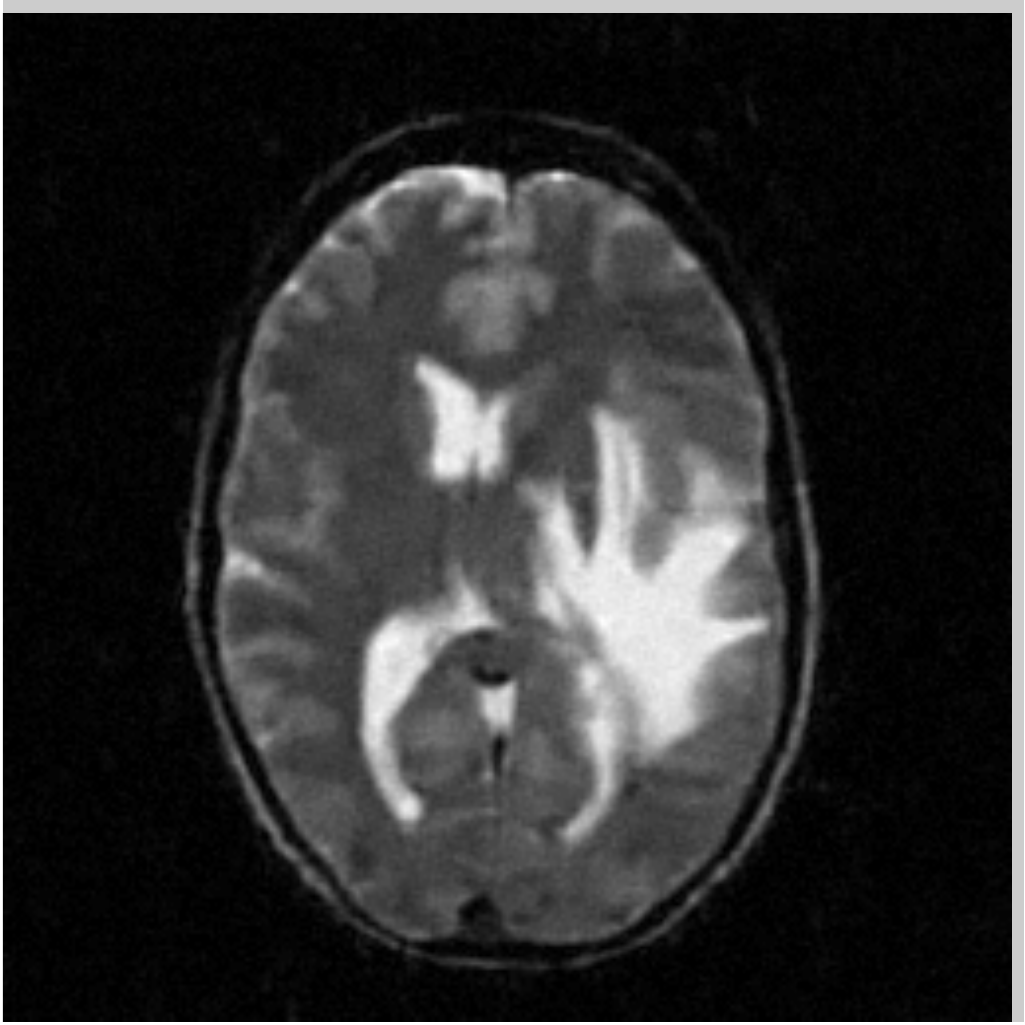}}
\subfigure[TV,SNR=17.43]{\includegraphics[width=0.23\textwidth,height=0.23\textwidth,trim=0.5cm 0.5cm 0.5cm 0.5cm,clip]{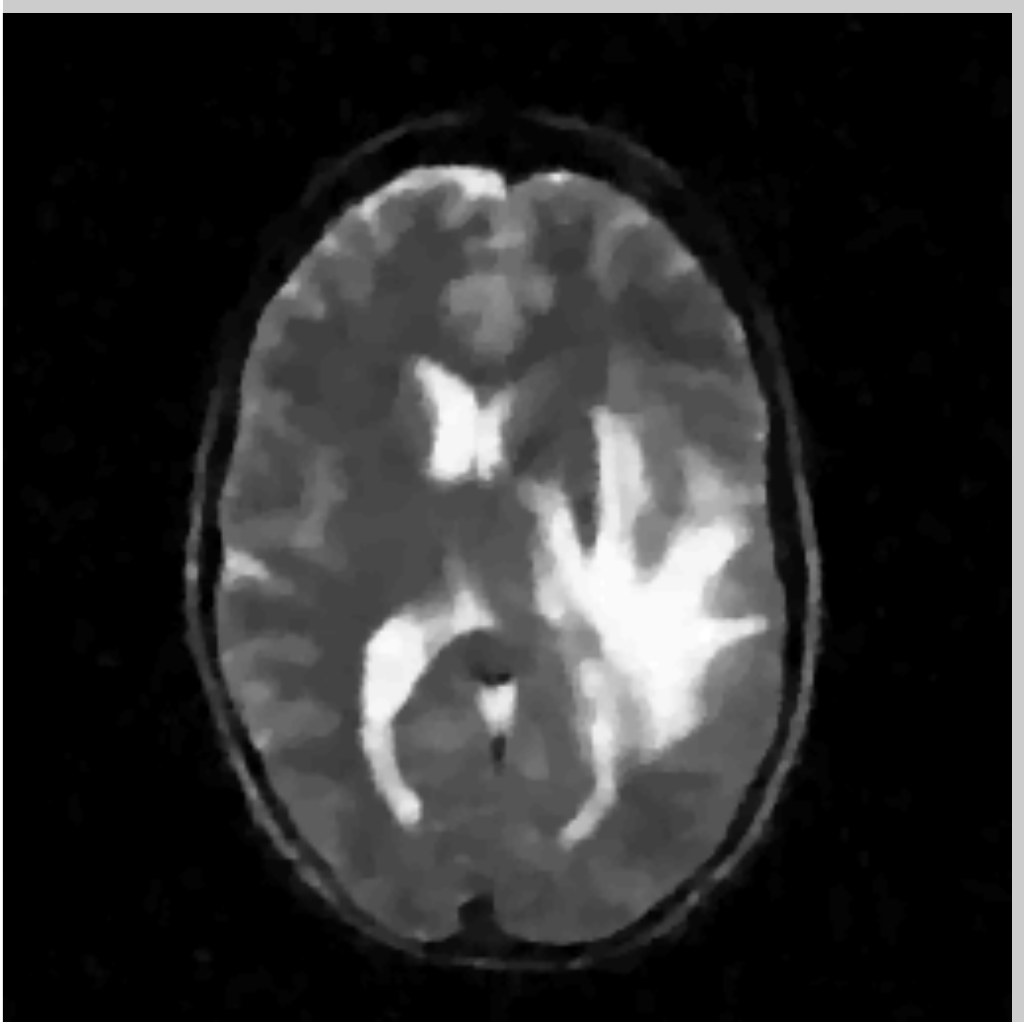}}
\subfigure[NLS,SNR=18.46]{\includegraphics[width=0.23\textwidth,height=0.23\textwidth,trim=0.5cm 0.5cm 0.5cm 0.5cm,clip]{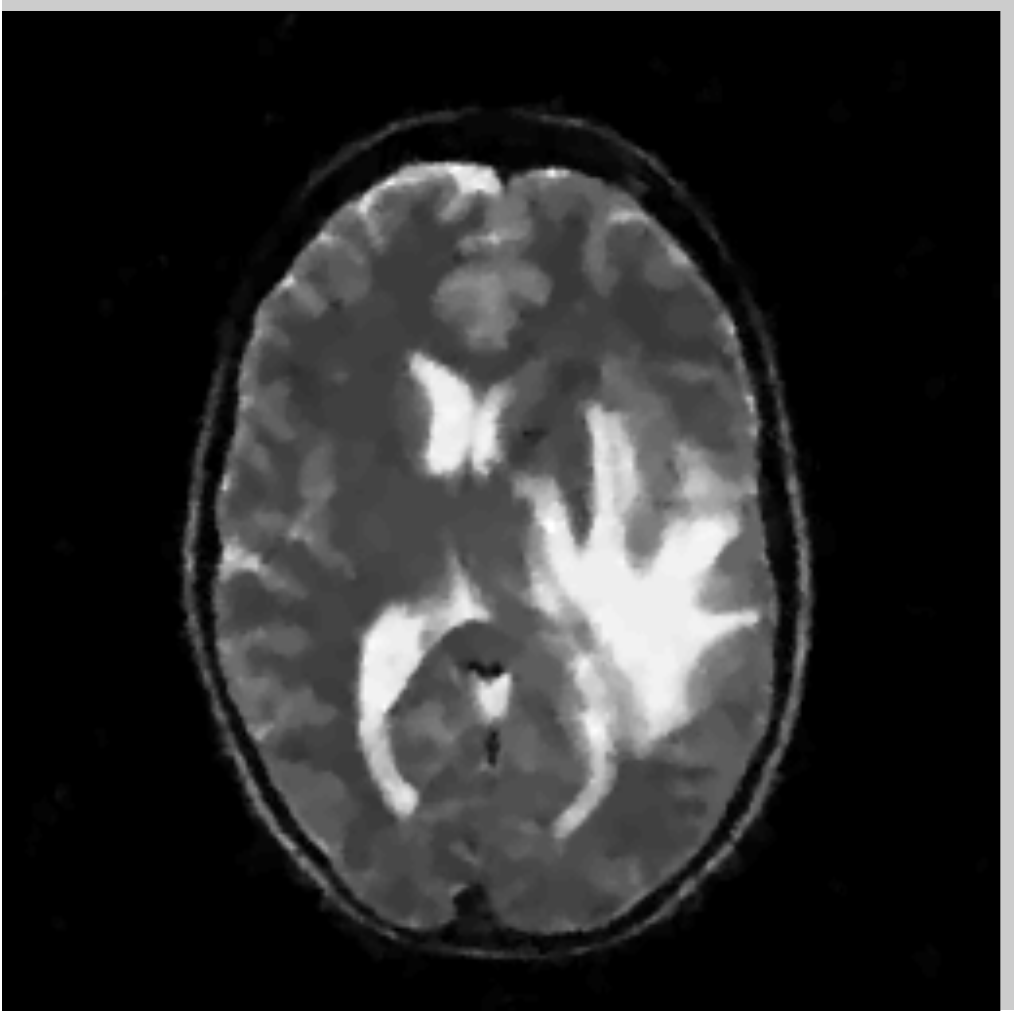}}
\\
\subfigure[Sampling pattern]{\includegraphics[width=0.23\textwidth,height=0.23\textwidth]{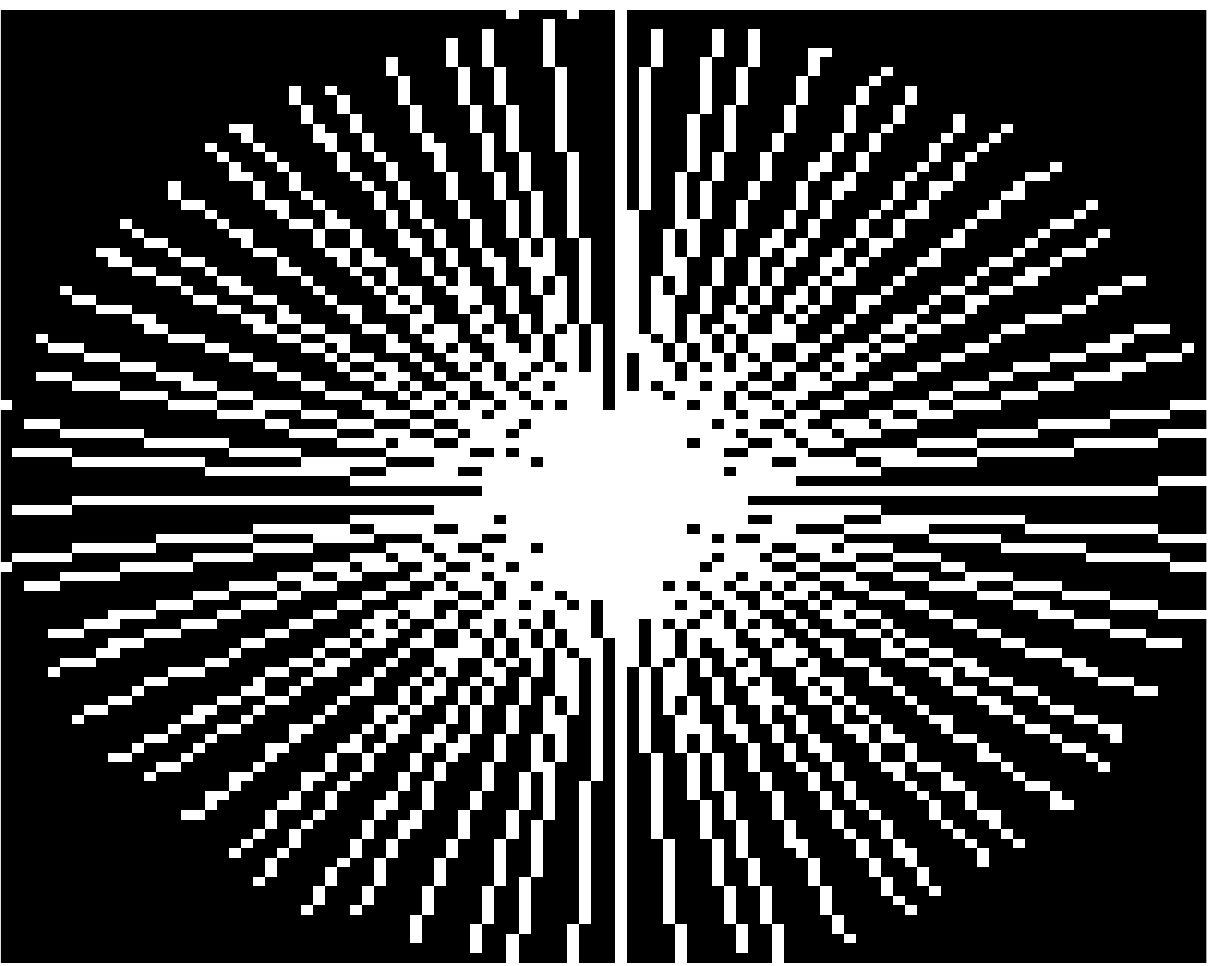}}
\subfigure[DLMRI error]{\includegraphics[width=0.23\textwidth,height=0.23\textwidth,trim=0.5cm 0.5cm 0.5cm 0.5cm,clip]{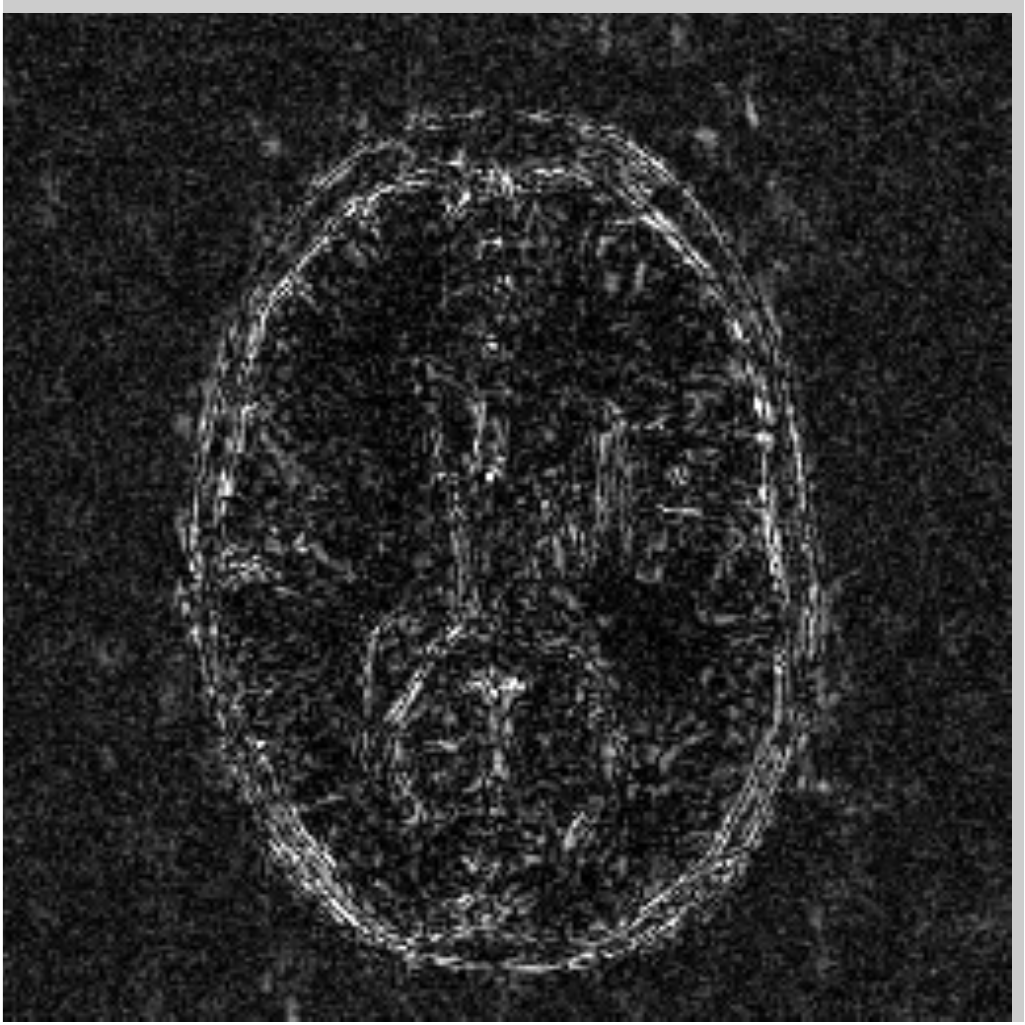}}
\subfigure[TV error]{\includegraphics[width=0.23\textwidth,height=0.23\textwidth,trim=0.5cm 0.5cm 0.5cm 0.5cm,clip]{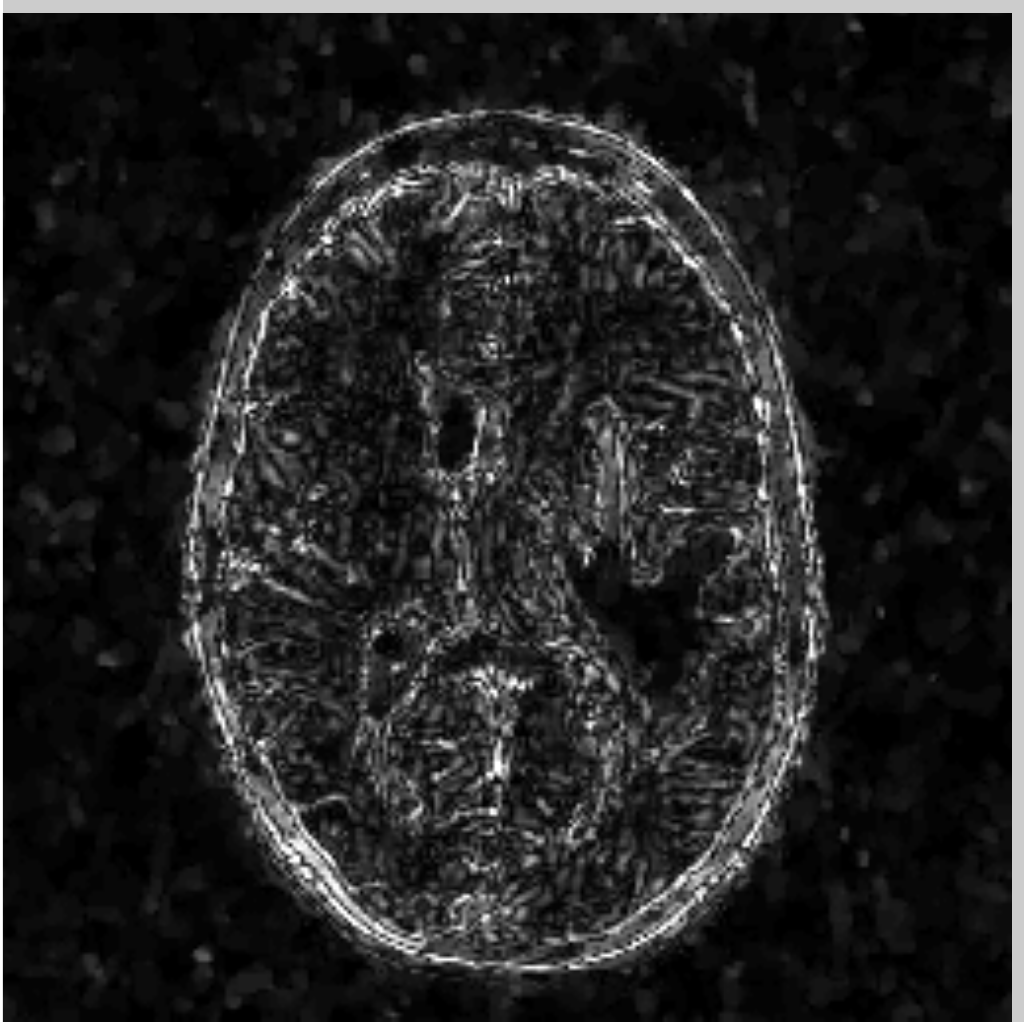}}
\subfigure[NLS error]{\includegraphics[width=0.23\textwidth,height=0.23\textwidth,trim=0.5cm 0.5cm 0.5cm 0.5cm,clip]{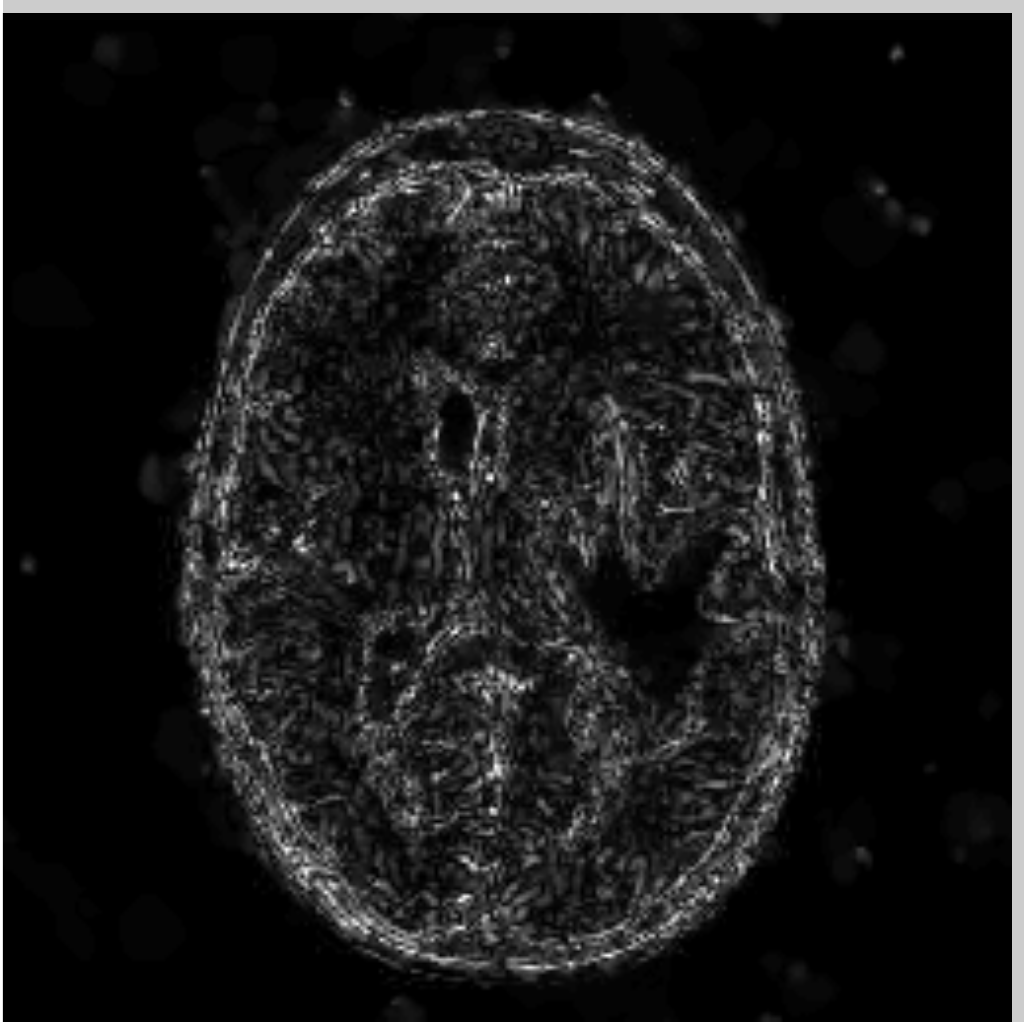}}

\caption{\small Comparison of the algorithms in the presence of noise. We consider the recovery of a $256\times 256$ original MRI brain image from its radial trajectory with 40 spokes, contaminated by Gaussian noise with standard deviation $\sigma=18.8$. The error images are magnified by a scale of 5 fold for the best visibility. This is a challenging case due to the high undersampling factor and high measurement noise. We observe that the NLS scheme provides the best overall reconstructions.}
\label{radial}
\end{figure*}

 The reconstructions of a $256\times 256$ brain image from its radial samples acquired with a  40 spoke trajectory are shown in Fig. \ref{radial}. The measurements are corrupted with zero mean complex Gaussian noise of standard deviation $\sigma=18.8$. All methods result in loss of subtle image features since the acceleration factor and the noise level are high. We observe that the NLS scheme provides better recovery than the competing methods. The quantitative results in this setting for various MR images are shown in the bottom section of Table IV. We observe that the SNR improvement offered by NLS over the other methods are not as high as in the previous cases, mainly due to the considerable noise in the data and the high acceleration.

 Finally, we show the recovery of four MR images from three fold radial under sampled data that is contaminated with zero mean complex Gaussian noise of standard deviation $\sigma=10$. These experiments show that the NLS scheme can be used to obtain good quality reconstructions at moderate acceleration factors and noise levels.

\begin{table}
\vspace{-2mm}
\centering
\label{noisyquantitative}
\begin{tabular}{c| c c | c c| c c }
\toprule
\bf Image &  \multicolumn{2}{c}{\bf DLMRI} & \multicolumn{2}{|c}{\bf TV} & \multicolumn{2}{|c}{\bf NLS}\\
\midrule
\hline
   & \small SNR   & \small PSNR    & \small SNR   & \small PSNR & \small SNR  & \small PSNR\\
\hline
\small Brain1  &13.55 & 22.82 & 14.81 & 24.67 &  \textbf{18.29} & \textbf{28.45}\\
\small Brain2  &14.38 & 24.74 & 16.10& 27.12& \textbf{18.63} & \textbf{29.83}\\
\small Brain3  & 13.10& 26.82& 16.19&30.37 & \textbf{19.73} &\textbf{33.80}\\
\small Ankle  &12.96&25.00 & 15.02 &27.64 & \textbf{18.52} &\textbf{31.13}\\
\small Spine & 16.33&26.57&18.38&29.29 &  \textbf{20.49} &\textbf{31.57}\\
\small Willis' & 14.56&26.35& 16.08&28.53 &  \textbf{18.14}&\textbf{30.45}\\
 \hline  \hline

\small Brain1  &\textbf{12.59}&\textbf{21.87} &11.84&21.27 &   12.35& 21.81\\
\small Brain2  &  17.46&27.83 & 17.43&28.14 & \textbf{18.46}&\textbf{29.54}\\
\small Brain3  & 14.13&27.85 & 16.98&31.00 &  \textbf{18.00}&\textbf{31.97}\\
\small Ankle  & 15.80&27.85 & 16.17&28.57 &  \textbf{16.81}& \textbf{29.57}\\
\small Spine & 18.54&28.77 &19.86&30.50 &   \textbf{20.30} &\textbf{30.82}\\
\small Willis'  & 14.18&25.97 & 14.55&26.69 &  \textbf{15.61}&\textbf{27.69}\\ \hline
\end{tabular}

\caption
{\small Quantitative comparison of the algorithms in the presence of noise. The top part shows the SNR of the reconstructions obtained from 4 fold Cartesian under sampled data, contaminated by zero mean complex Gaussian noise with standard deviation $\sigma=10$. The bottom part shows the SNR of the reconstructions from radial under sampled data with 40 spokes, contaminated by zero mean complex Gaussian noise with standard deviation $\sigma=18.8$. The quantitative results show that the proposed iterative NLS scheme provides consistently better reconstructions for the above cases.}

\end{table}
\section{Conclusion}
We introduced a fast iterative non-local shrinkage algorithm to recover MR image data from under sampled Fourier measurements. This approach is enabled by the reformulation of current non-local schemes as an iterative re-weighting algorithm to minimize a global criterion \cite{Wendy}. The proposed algorithm alternates between a non-local shrinkage step and a quadratic subproblem, which can be solved analytically and efficiently. We derived analytical shrinkage rules for several penalties that are relevant in non-local regularization. We accelerated the non-local shrinkage step, whose direct evaluation involves expensive non-local patch comparisons, by exploiting the redundancy between the terms at adjacent pixels. The resulting algorithm is observed to be considerably faster than our previous implementation. The comparison of different penalties demonstrated the benefit in using distance functions that saturate with distant patches. The comparisons of the proposed scheme with state of the art algorithms show a considerable reduction in alias artifacts and preservation of edges.

\section*{Appendix A: Simplification of Eq (\ref{newf})}
Using the formula for the shrinkage from (\ref{pshrinkage}), specified by $\mbf s_{\mbf x,\mbf q} = \bkt{P_{\mbf x} (\mbf f) - P_{\mbf x + \mbf q} (\mbf f)}\nu\bkt{\|P_{\mbf x} (\mbf f) - P_{\mbf x + \mbf q} (\mbf f)\|}$, we obtain
\begin{equation}
{\cal R}(\mbf f) = \sum_{\mathbf{x},} \sum_{\mathbf{q}}\|P_{\mathbf{x}}(\mbf f)-P_{\mathbf{x+q}}(\mbf f)-\mbf s_{\mbf x,\mbf q}\|^2
\end{equation}
Expanding the above expression
\begin{equation}
\label{newreg}
{\cal R}(\mbf f)=\sum_{\mathbf{x},} \sum_{\mathbf{q}} \sum_{\mathbf{p}\in{\cal B}}\abs{f(\mathbf{x}+\mathbf{p})-f(\mathbf{x+q}+\mathbf{p}) -\mbf s_{\mbf x,\mbf q}(\mathbf{p})}^{2}
\end{equation}
We use a change of variables $\mbf x = \mbf x+\mbf p$ to obtain
\begin{eqnarray}\nonumber
{\cal R}(\mbf f) &=& \sum_{\mathbf{x},} \sum_{\mathbf{q}} \sum_{\mathbf{p}\in{\cal B}}\abs{\overbrace{f(\mbf x)-f(\mbf x+\mbf q)}^{e\bkt{\mbf x,\mbf q}} -\mbf s_{\mbf x-\mbf p,\mbf q}(\mathbf{p})}^{2}\\
 &=& \sum_{\mathbf{x},\mathbf{y}} \abs{e\bkt{\mbf x,\mbf q}}^{2}+ c\\\nonumber&&~ + 2 \sum_{\mathbf{x},\mathbf{y}} e\bkt{\mbf x,\mbf q}\underbrace{\sum_{\mathbf{p}\in{\cal B}}{\mbf s_{\mbf x-\mbf p,\mbf q}(\mathbf{p})}}_{h_{\mbf q}(\mathbf x)} \\\nonumber\label{consolidated}
&=&\sum_{\mathbf{x}}\sum_{\mathbf y\in \mbf x + {\cal N}}\abs{f(\mathbf x)-f(\mathbf y)- h_{\mbf q}\bkt{\mbf x}}^{2} + c - d.\\
\end{eqnarray}
In the above equations, $c$ and $d$ are constants specified by
\begin{eqnarray*}
c &=& \sum_{\mathbf{x},\mathbf{y}} \sum_{\mathbf{p}\in{\cal B}} \abs{\mbf s_{\mbf x-\mbf p,\mbf q}(\mathbf{p})}^{2}\\
d &=& \sum_{\mathbf{x},\mathbf{y}} \abs{h_{\mbf q}(\mbf x)}^{2}.
\end{eqnarray*}

Since the solution to (\ref{newf}) does not depend on the constants, we ignore these terms. Thus, (\ref{newreg}) can be rewritten using (\ref{consolidated}) as
\begin{eqnarray*}
{\cal R}(\mbf f) &=&\sum_{\mathbf q\in {\cal N}}\|\underbrace{f(\mathbf x)-f(\mathbf x + \mbf q)}_{{\cal D}_{\mbf q}f}- h_{\mbf q}(\mbf x)\|^{2}
\end{eqnarray*}
Here, ${\cal D}_{\mathbf{q}} f(\mathbf{x}) = f(\mathbf{x}+\mathbf{q})-f(\mathbf{x})$ is the finite difference operator.

We observe that the expression for $h_{\mbf q}(\mbf x)$
\begin{equation}
\label{hq2}
h_{\mbf q}(\mathbf x) = \sum_{\mathbf{p}\in{\cal B}}{\mbf s_{\mbf x-\mbf p,\mbf q}(\mathbf{p})},\end{equation}
 can be further simplified. From (\ref{pshrinkage}), we have the patch $\mbf s$ specified as
\begin{eqnarray*}
\mbf s_{\mbf x,\mbf q}&=&\bkt{P_{\mbf x}{f} - P_{\mbf x+\mbf q}{f}}~\underbrace{\nu\bkt{\|P_{\mbf x}{f} - P_{\mbf x+\mbf q}{f}\|}}_{u_{\mbf q}(\mbf x)}
\end{eqnarray*}
Here, $u_{\mbf q}(\mbf x) =\nu\bkt{\|P_{\mbf x}f - P_{\mbf x+\mbf q}f\|}$ is the factor between 0 and 1, which is multiplied by the patch to get the shrinked patch. Hence, $\mbf s_{\mbf x,\mbf q}(\mbf r) =\left[f(\mbf x+\mbf r) - f(\mbf x+\mbf q+\mbf r)\right]\cdot u_{\mbf q}(\mbf x); \mbf r\in \cal B$. Thus, we have
\begin{equation}
\mbf s_{\mbf x-\mbf p,\mbf q}(\mathbf{p}) = \left[f(\mbf x) - f(\mbf x+\mbf q)\right]~u_{\mbf q}(\mbf x-\mbf p).
\end{equation}
Substituting in (\ref{hq2}), we get
\begin{equation}
h_{\mbf q}(\mathbf x) =\bkt{f(\mbf x) - f(\mbf x+\mbf q)} \underbrace{\sum_{\mathbf{p}\in{\cal B}}u_{\mbf q}(\mbf x-\mbf p)}_{v_{\mbf q}(\mbf x)}.
\end{equation}

\section*{Appendix B: Shrinkage rules for useful non-local distance functions}
\subsubsection{Thresholded $\ell_{p};p\leq 1$ metric}
We now consider the saturating $\ell_{p}$ metric, specified by
\begin{equation}
\label{distance}
\phi(t) = \left\{\begin{array}{ccc}
|t|^{p}/p & \mbox{ if } & |t| < T\\
T^{p}/p & \mbox{else}.
\end{array}
\right.
\end{equation}

Computing the shrinkage rule for this mapping according to (\ref{combined}), we obtain
\begin{equation}
\nu(t) = \left\{\begin{array}{ccc}
0 & \mbox{ if } & |t| < \beta^{1/(p-2)}\\
1 - \frac{1}{\beta}|t|^{p-2} & \mbox{ if } &\beta^{1/(p-2)} \leq |t| < T\\
1  & \mbox{else},
\end{array}
\right.
\end{equation}

Additionally, taking $T = \infty$ we get the shrinkage rule for the unthresholded $\ell_p$ metric as
\begin{equation}
\nu(t) = \left\{\begin{array}{ccc}
0 & \mbox{ if } & |t| < \beta^{1/(p-2)}\\
1 - \frac{1}{\beta}|t|^{p-2} & \mbox{ else }
\end{array}
\right.
\end{equation}

\begin{figure*}
\vspace{-7mm}
\centering
\subfigure[Circle of Willis]{\includegraphics[width=3.5cm,height=3.5cm]{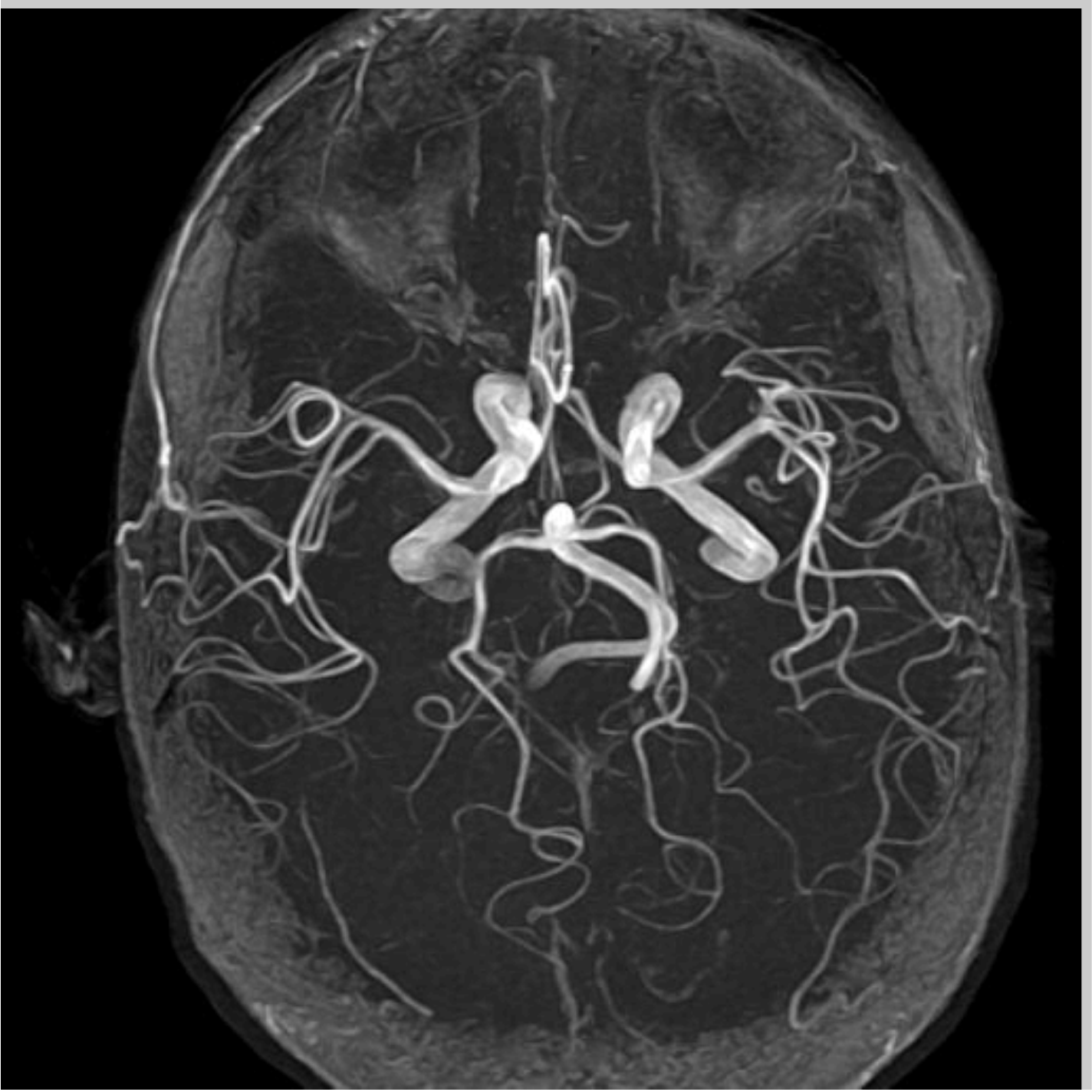}}
\subfigure[Spine]{\includegraphics[width=3.5cm,height=3.5cm]{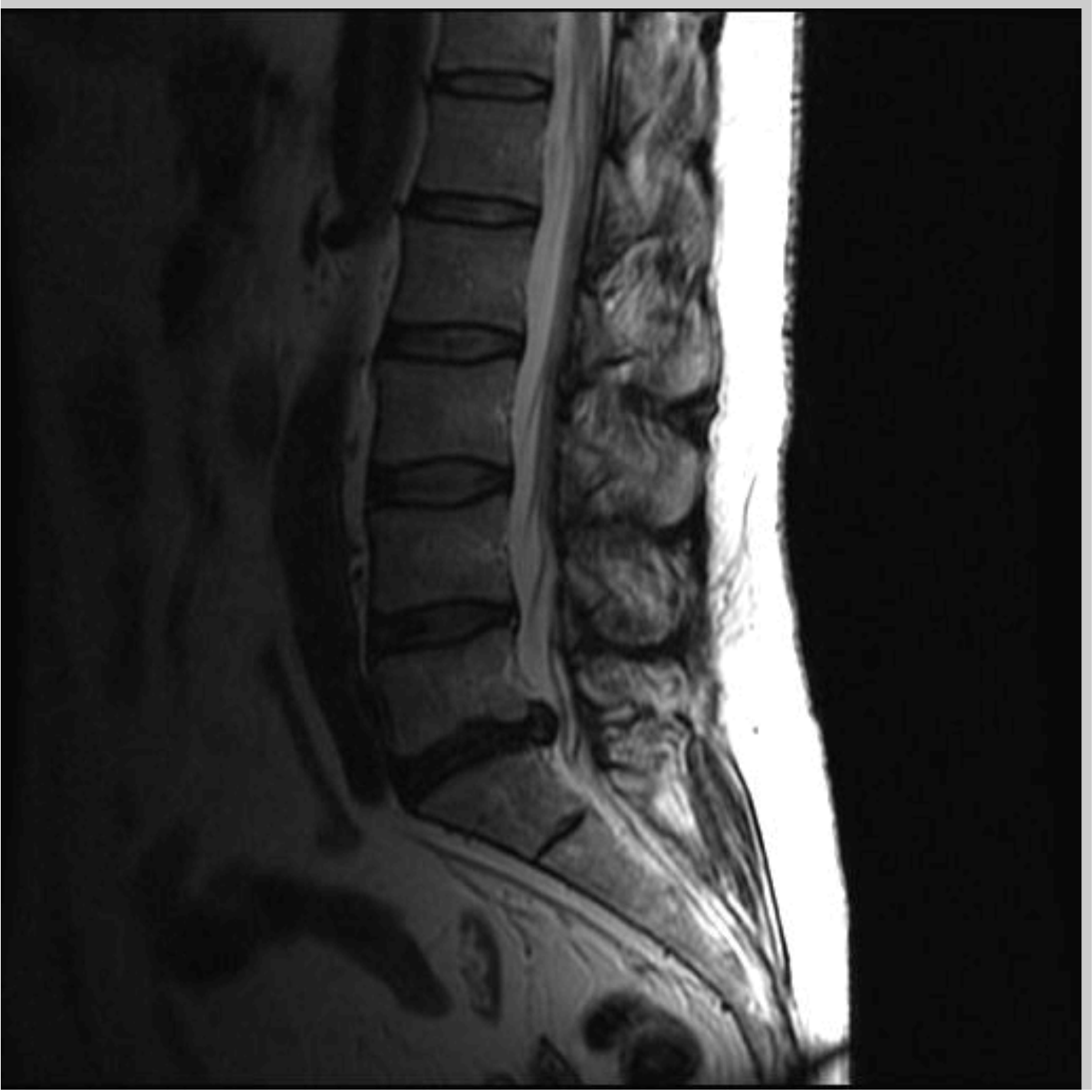}}
\subfigure[Brain3]{\includegraphics[width=3.5cm,height=3.5cm]{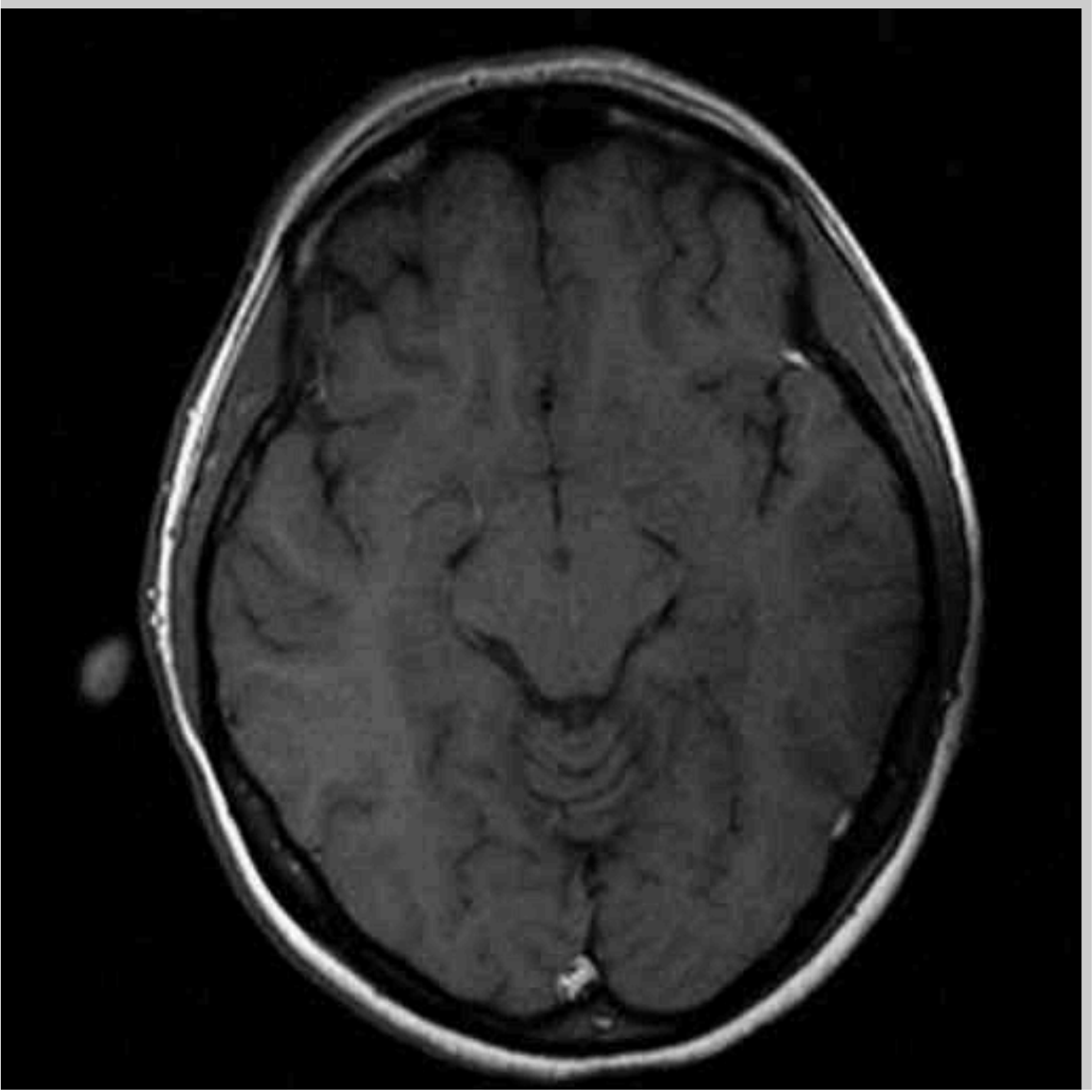}}
\subfigure[Brain4]{\includegraphics[width=3.5cm,height=3.5cm]{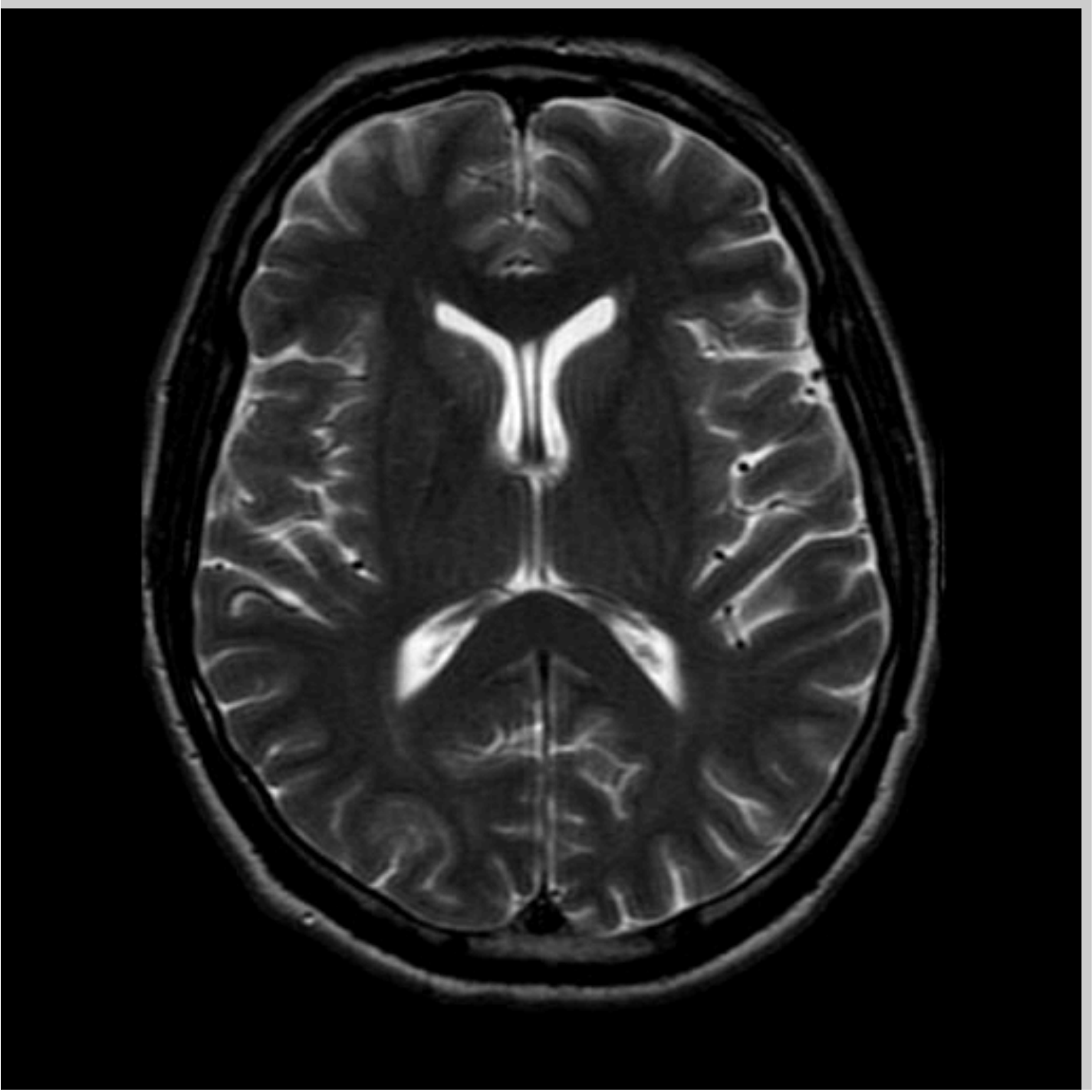}}\\

\subfigure[NLS,SNR=24.01]{\includegraphics[width=3.5cm,height=3.5cm]{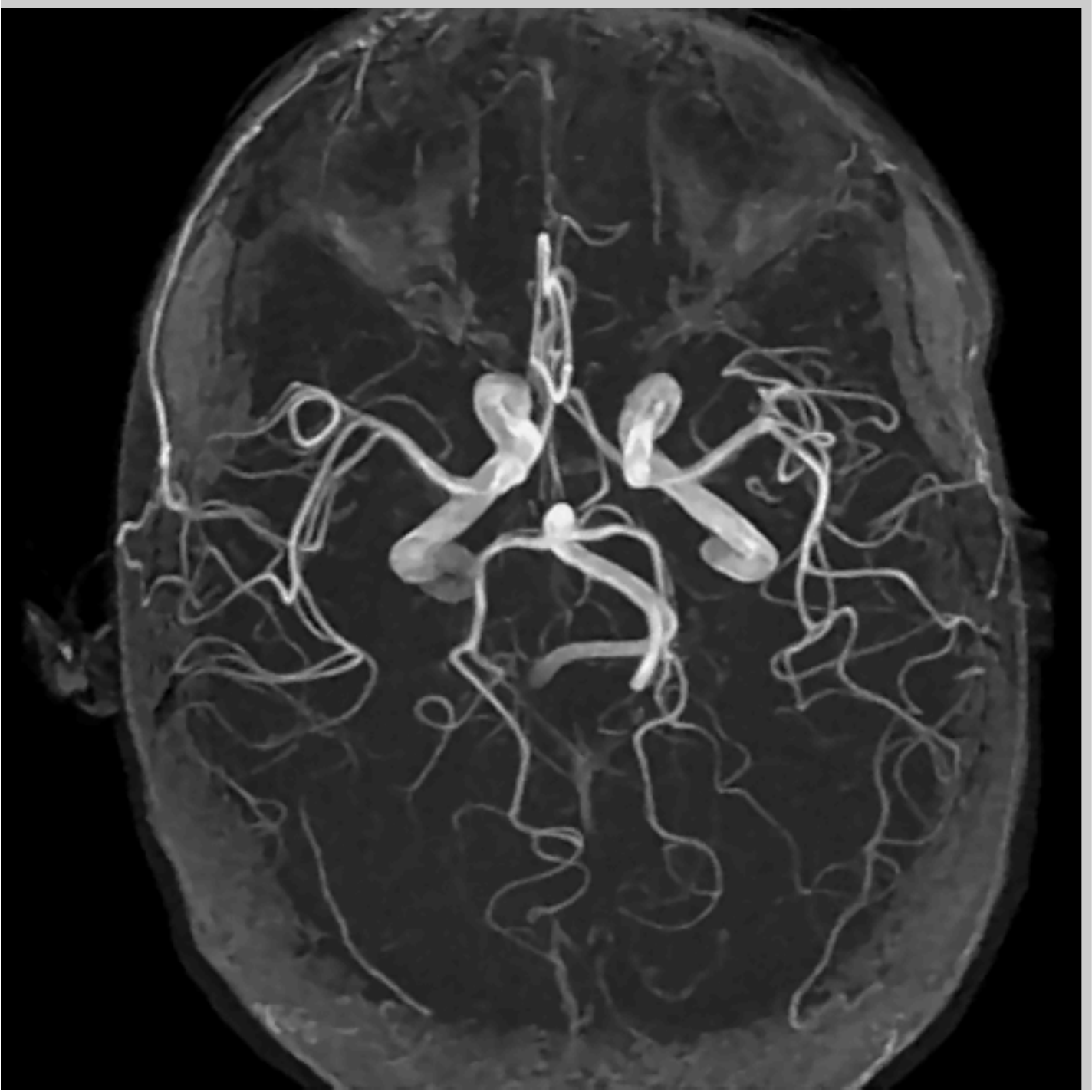}}
\subfigure[NLS,SNR=29.43]{\includegraphics[width=3.5cm,height=3.5cm]{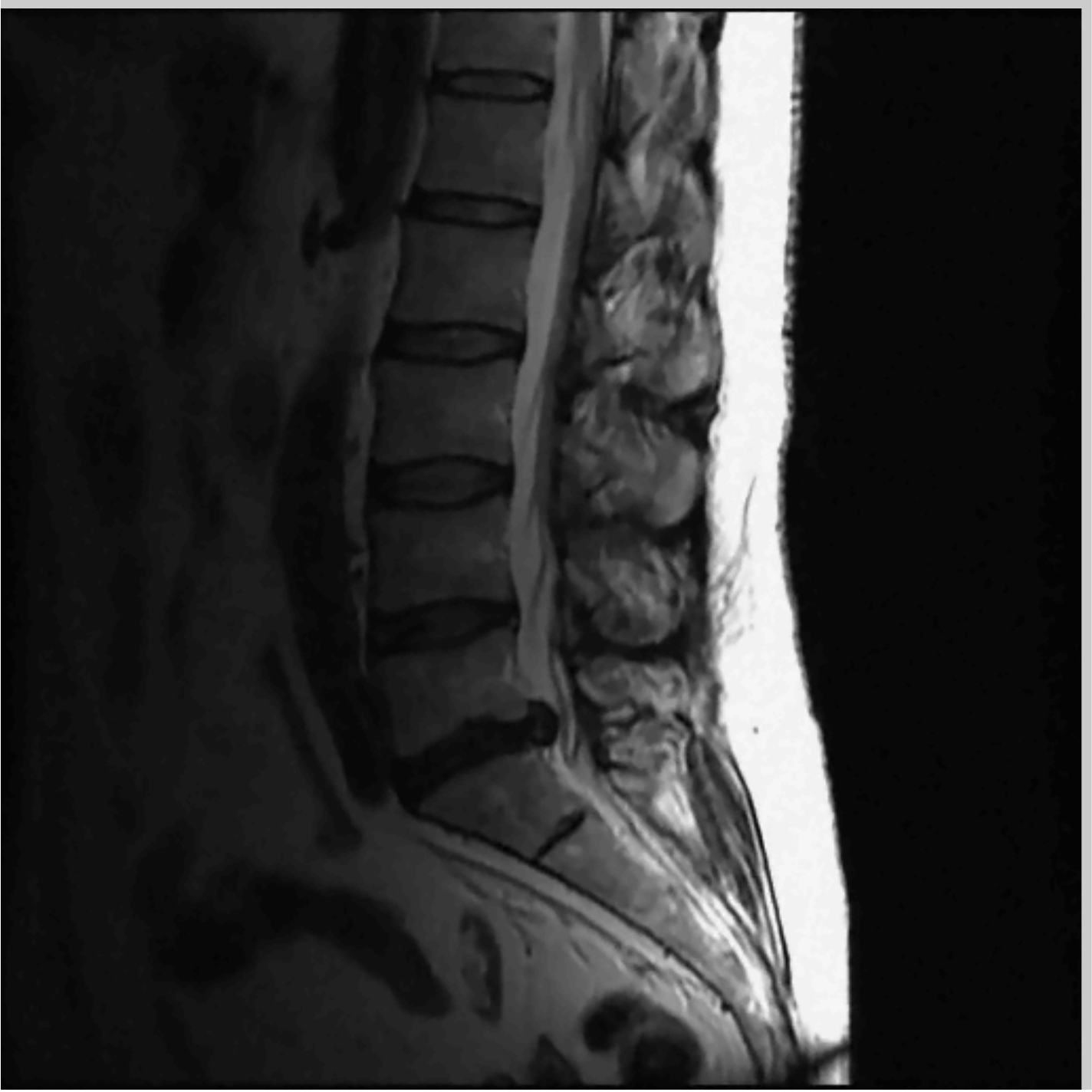}}
\subfigure[NLS,SNR=25.65]{\includegraphics[width=3.5cm,height=3.5cm]{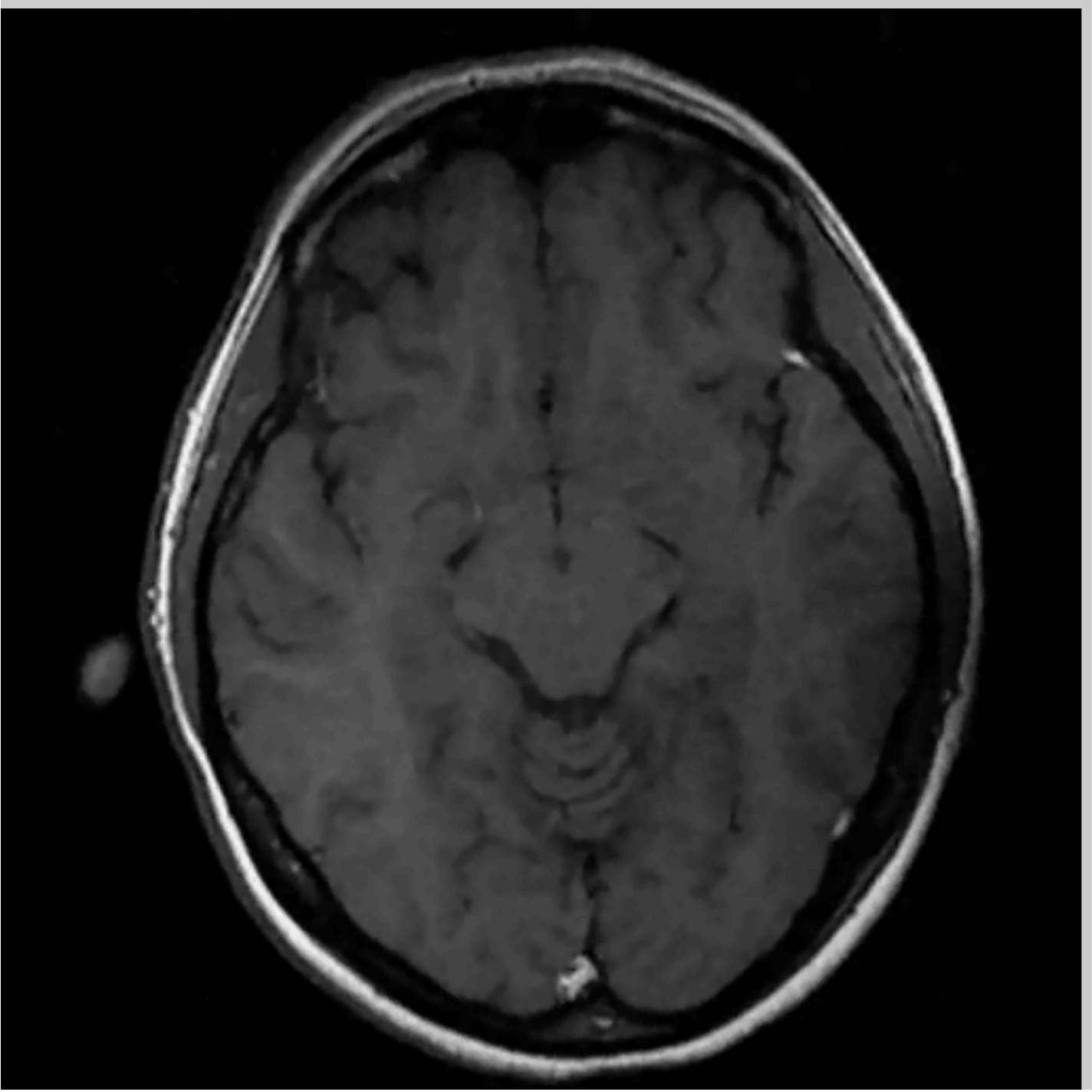}}
\subfigure[NLS,SNR=24.75]{\includegraphics[width=3.5cm,height=3.5cm]{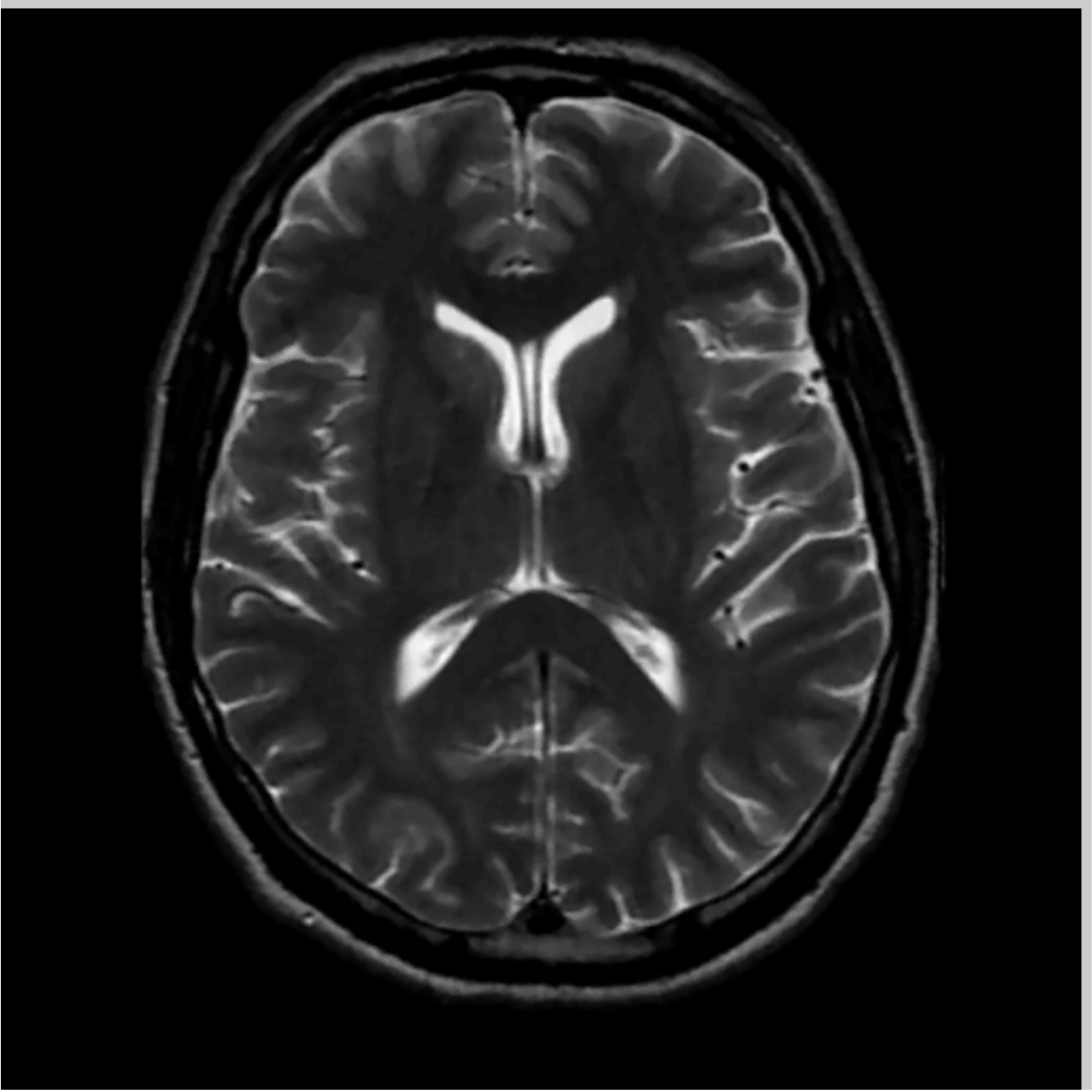}}\\

\subfigure[Error]{\includegraphics[width=3.5cm,height=3.5cm]{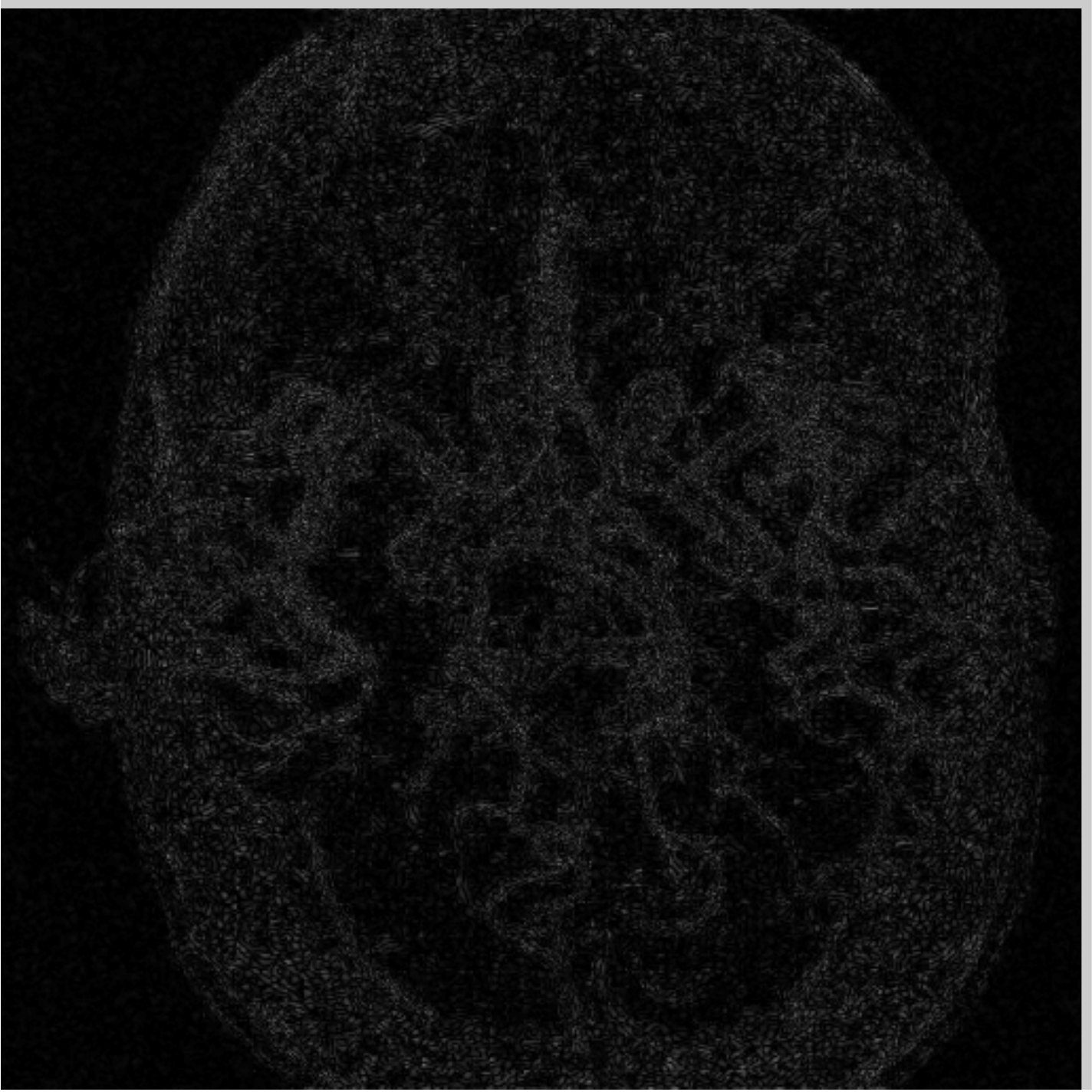}}
\subfigure[Error]{\includegraphics[width=3.5cm,height=3.5cm]{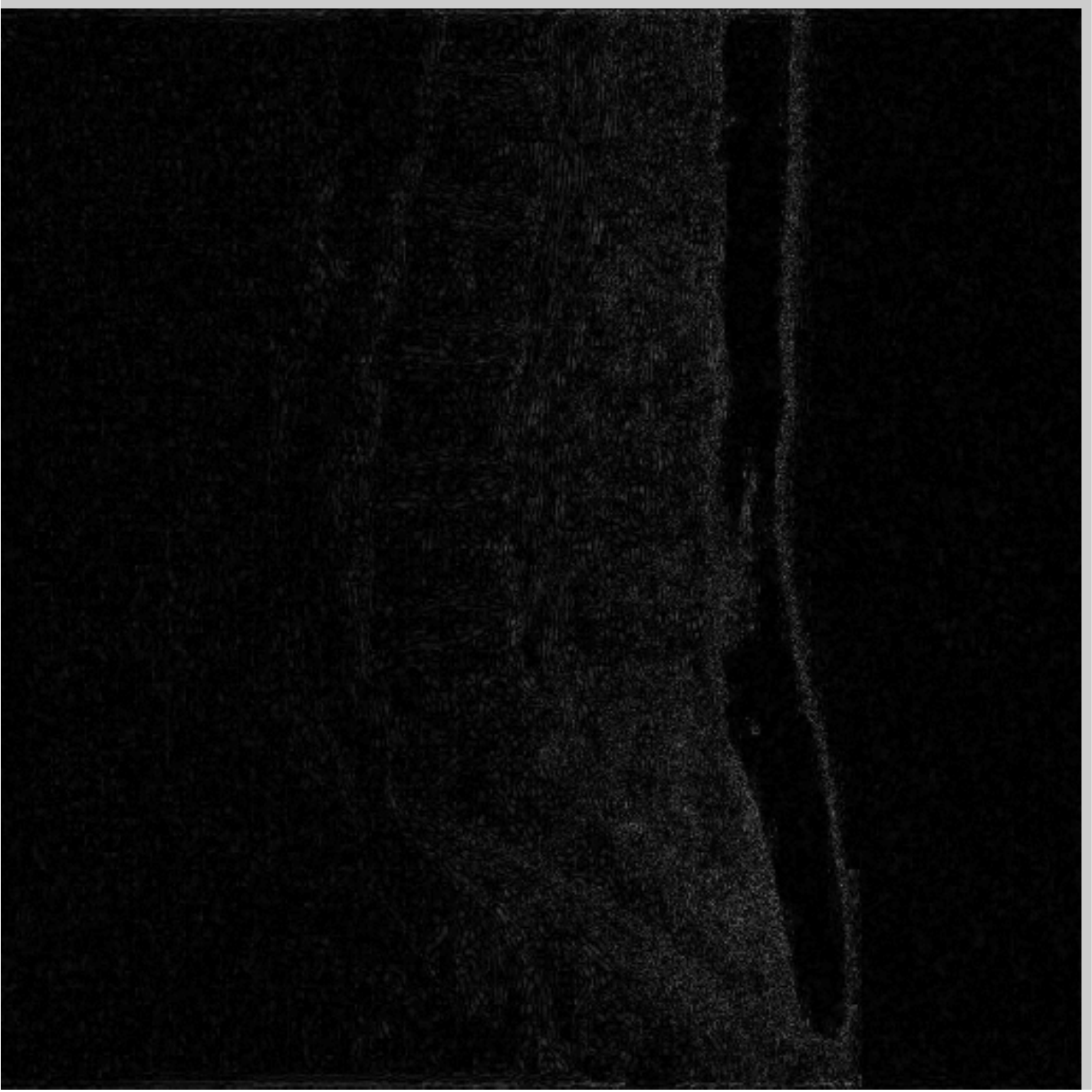}}
\subfigure[Error]{\includegraphics[width=3.5cm,height=3.5cm]{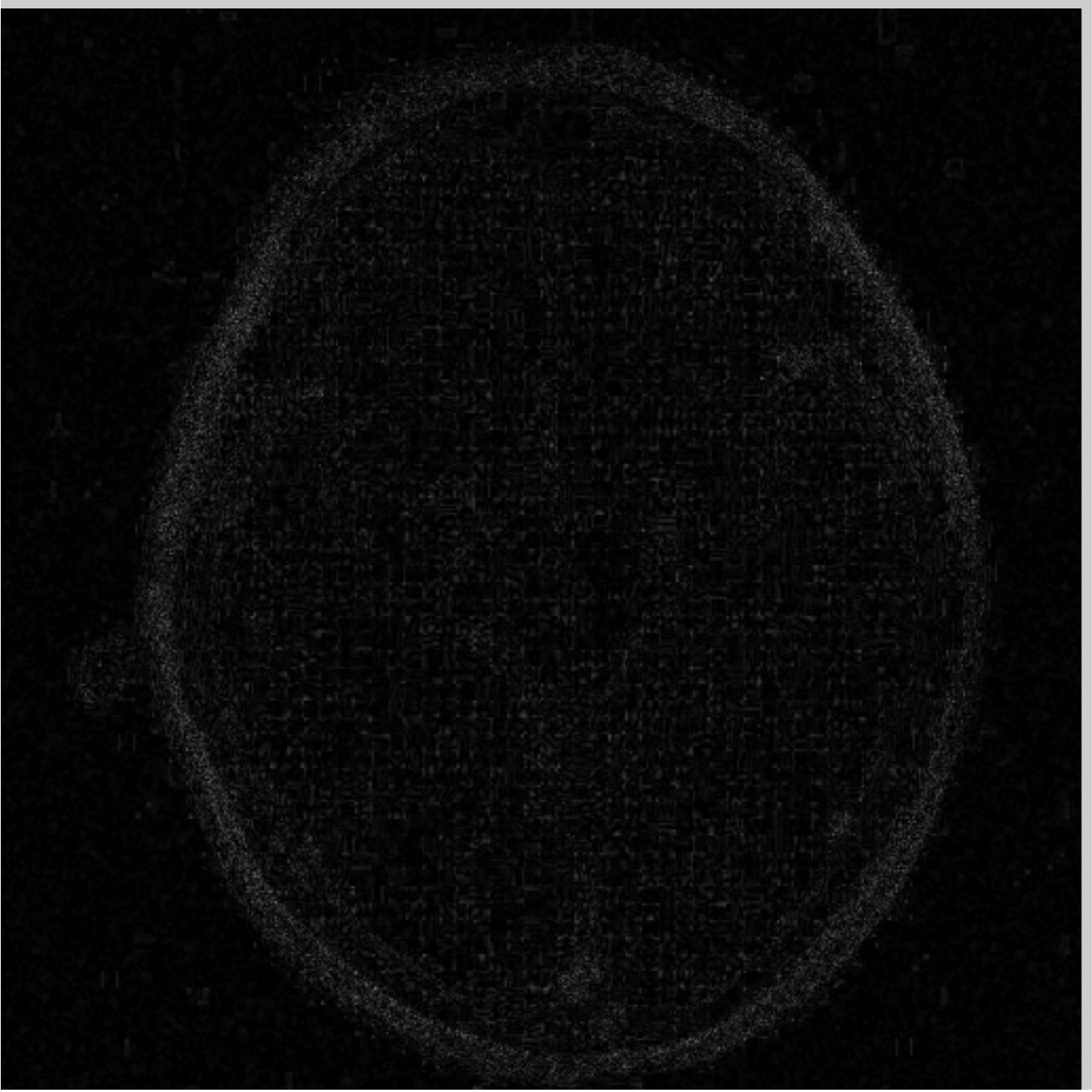}}
\subfigure[Error]{\includegraphics[width=3.5cm,height=3.5cm]{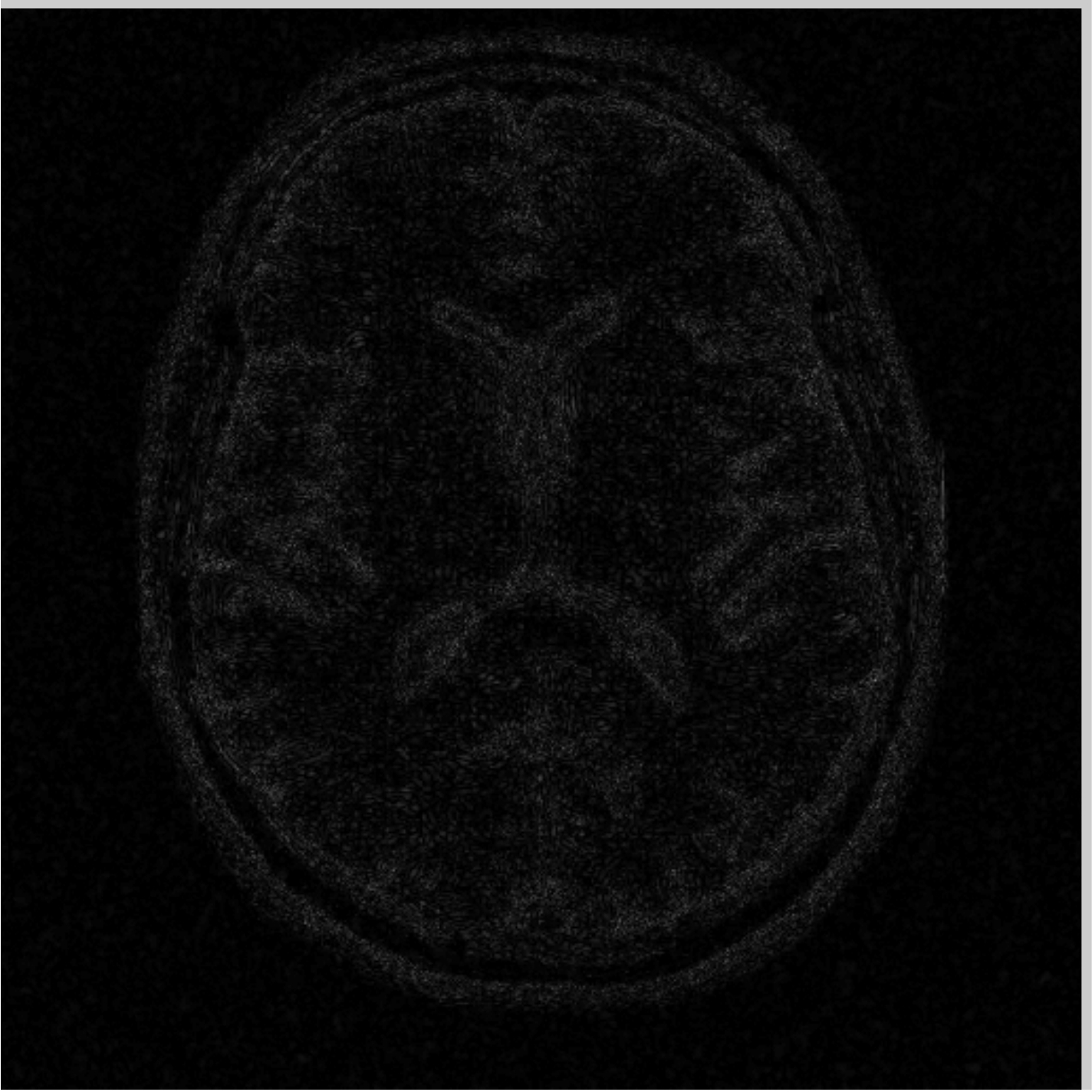}}\\
\caption{\small Comparison of different MR images using NLS algorithms in the presence of noise. We consider the recovery from a three fold undersampled radial sampling pattern, contaminated by zero mean complex Gaussian noise with standard deviation $\sigma=10$. The top two rows show the original and reconstructed images, while the error images scale by a factor of five are shown in the bottom row. We observe that the NLS scheme preserves well the edges and the fine details at low acceleration with presence of noise.}
\label{NLSfinaladded}
\end{figure*}
\subsubsection{Penalty corresponding to alternating $H_1$ non-local scheme}
We now consider the $H_1$ metric, specified by
\begin{equation}
\varphi(t) = 1-\exp\bkt{-\frac{t^{2}}{2\sigma^{2}}}
\end{equation}
Computing the shrinkage rule, we obtain
\begin{equation}
\nu(t) = \left\{\begin{array}{ccc}
0  &\mbox{  if  }& \exp\bkt{-\frac{t^{2}}{2\sigma^{2}}}  > \beta\sigma^{2}\\
1-\frac{\exp\bkt{-\frac{t^{2}}{2\sigma^{2}}}}{\beta\sigma^{2}} & \mbox{ else }
\end{array}
\right.
\end{equation}
\subsubsection{Penalty corresponding to Peyre's non-local scheme}
We now consider the penalty corresponding to Peyre's alternating scheme \cite{Wendy,peyre2011}:
\begin{equation}
\phi(t) = 1-\exp\bkt{-\frac{t}{\sigma}}
\end{equation}
Computing the shrinkage rule, we obtain
\begin{equation}
\nu(t) = \left\{\begin{array}{ccc}
0  &\mbox{  if  }& \exp\bkt{-\frac{t}{\sigma}}  > \beta\sigma t\\
1-\frac{\exp\bkt{-\frac{t}{\sigma}}}{\beta\sigma t} & \mbox{ else }
\end{array}
\right.
\end{equation}
\subsubsection{Penalty corresponding to alternating non-local TV scheme}
The penalty function for the alternating non-local TV scheme is specified by \cite{Wendy,YLou}:
\begin{equation}
\phi(t) = {\rm erf}\bkt{\frac{t}{
\sigma}}\end{equation}
Computing the shrinkage rule, we obtain
\begin{equation}
\nu(t) = \left\{\begin{array}{ccc}
0  &\mbox{  if  }& \frac{2}{\sqrt{\pi}}\exp\bkt{-\frac{t^2}{\sigma^2}}  > \beta\sigma t\\
1-\frac{2}{\sqrt{\pi}}\frac{\exp\bkt{-\frac{t^2}{\sigma^2}}}{\beta\sigma t} & \mbox{ else }
\end{array}
\right.
\end{equation}

\bibliographystyle{IEEEbib}
\bibliography{NLM2}

\end{document}